\pdfoutput=1

\documentclass[11pt]{article}

\usepackage[]{acl}

\usepackage{times}
\usepackage{latexsym}

\usepackage[T1]{fontenc}

\usepackage[utf8]{inputenc}

\usepackage{microtype}

\usepackage{graphicx}
\usepackage{amsmath}
\usepackage{amsfonts}
\usepackage{array,booktabs,arydshln,xcolor}
\usepackage{graphics}
 \usepackage[most]{tcolorbox}
\usepackage{lipsum}
\usepackage{pifont}
\usepackage{hyperref}
\usepackage{makecell}
\usepackage[super]{nth}
\usepackage{wasysym}
\usepackage{float}
\usepackage{resizegather}
\usepackage{adjustbox}
\usepackage{booktabs}
\usepackage{multirow}
\usepackage{makecell}
\usepackage{subcaption}
\usepackage{bm}

\usepackage{enumitem}

\newcommand{\removedbyraph}[1]{}

\title{Conversation Style Transfer using Few-Shot Learning}

\author{
    Shamik Roy\Thanks{~Correspondence to \texttt{royshami@amazon.com}}\quad
    Raphael Shu\quad 
    Nikolaos Pappas\quad
    Elman Mansimov\quad
    \\
    \textbf{
    Yi Zhang\quad
    Saab Mansour\quad
    Dan Roth
    }
    \\
    AWS AI Labs\\
}

\begin{document}
\maketitle
\begin{abstract}
Conventional text style transfer approaches focus on sentence-level style transfer without considering contextual information, and the style is described with attributes (e.g., formality). When applying style transfer in conversations such as task-oriented dialogues, existing approaches suffer from these limitations as context can play an important role and the style attributes are often difficult to define in conversations. In this paper, we introduce conversation style transfer as a few-shot learning problem, where the model learns to perform style transfer by observing only a few example dialogues in the target style. We propose a novel in-context learning approach to solve the task with style-free dialogues as a pivot. Human evaluation shows that by incorporating multi-turn context, the model is able to match the target style while having better appropriateness and semantic correctness compared to utterance/sentence-level style transfer. Additionally, we show that conversation style transfer can also benefit downstream tasks. For example, in multi-domain intent classification tasks, the F1 scores improve after transferring the style of training data to match the style of the test data.


\end{abstract}

\section{Introduction}\label{sec:introduction}
Recent advances in neural dialogue models \citep{gao2018neural,zhang2020recent,ni2022recent} enabled the handling of complex conversational scenarios. However, one key challenge that still remains in conversational AI is to obtain the desired conversation style. Conversations in nature are dynamic and the style requirement of utterances in a conversation depends on many factors including domain (e.g., banking vs restaurant), situation (e.g., conversation with someone angry vs depressed), the speaker demographics (e.g., senior vs youngster) among others, making style transfer of the whole conversation more challenging compared to style transfer of a single utterance. 



\begin{figure}[t!]
  \includegraphics[width=0.47\textwidth]{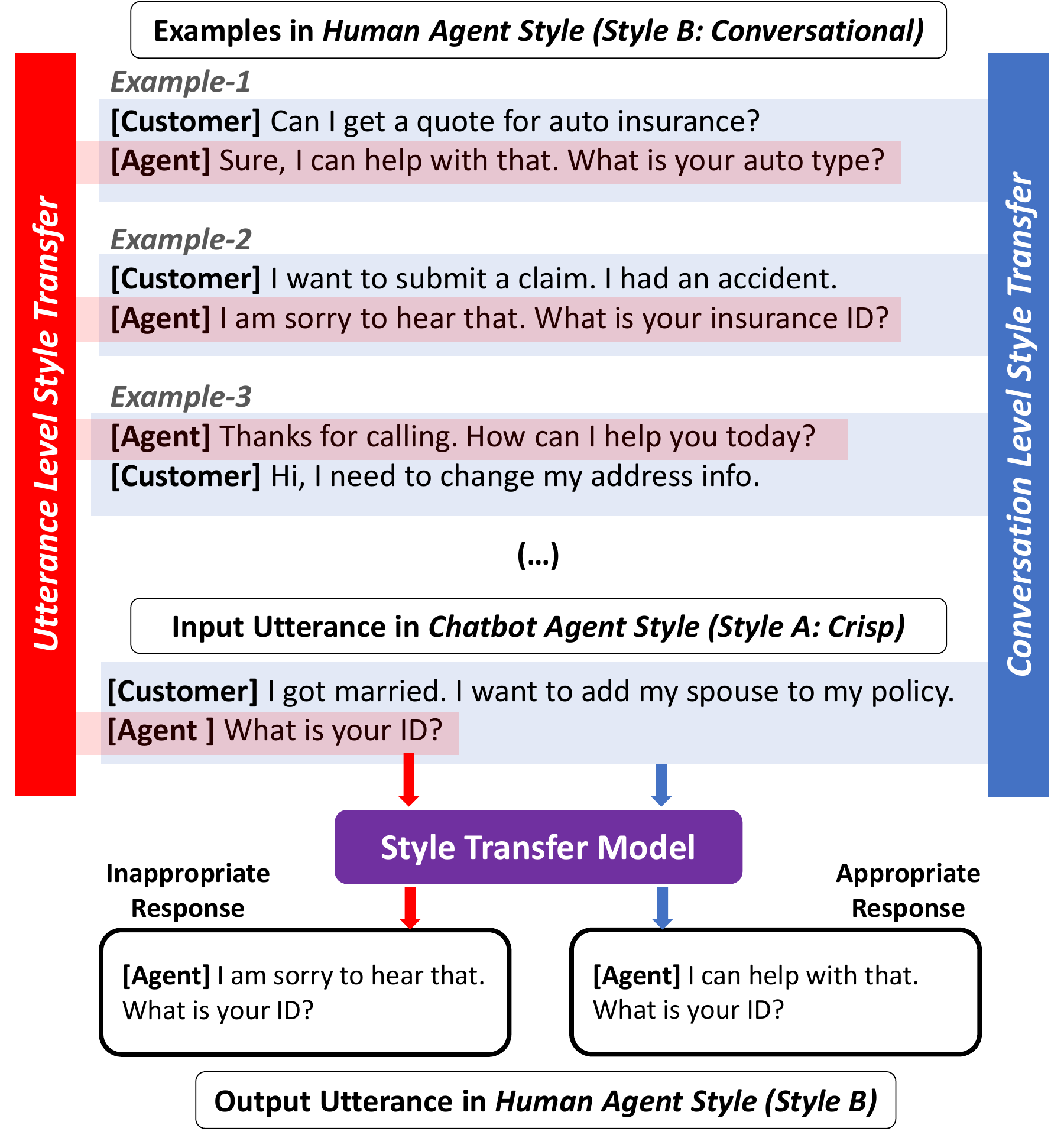}
  \caption{Transferring the style of an utterance from chatbot to human agent style based on three utterance-level (single-turn) and conversation-level (multi-turn) examples as input. Additional conversational context helps the style transfer model to yield a more appropriate response as the dialogue contains useful information that can be leveraged during the generation process.} 
  
  \label{fig:problem-formulation}
\end{figure}

Existing studies on Text Style Transfer (TST) focus on transferring the style at the sentence level from one known style to another \citep{pavlick2016empirical,rao2018dear,niu2018multi,wang2019harnessing,briakou2021ola} by ignoring contextual information, such as the previous turns in a conversation. However, as demonstrated in Figure \ref{fig:problem-formulation}, the context plays an important role in defining conversation style.



In this paper, we explore a novel task: \textit{few-shot learning for conversation style transfer}. Here, a style transfer model is expected to convert the style of an input conversation based on a few example conversations in the target style. This is in contrast with the common methodologies in TST, where the style is assumed to be describable with known and well-defined attributes (e.g., politeness, friendliness) \citep{zhang2018style,madaan2020politeness,reif2022recipe}. For conversations, defining such attributes is challenging due to the dynamic nature and domain dependency. Also, the style of a conversation may be a combination of many attributes. Examples from the TWCS dataset~\cite{stuart_axelbrooke_2017} in Table \ref{tab:off-the-shelf-example} show that the agent responses from Chipotle and VirginTrains services are identified to have similar politeness scores by an off-the-shelf politeness classifier \cite{danescu2013computational}, however, their actual styles are drastically different upon a closer look. 



\begin{table}
\centering
\resizebox{1\columnwidth}{!}{%
\begin{tabular}
{>{\arraybackslash}m{8.3cm}>{\centering\arraybackslash}m{2cm}>{\centering\arraybackslash}m{2.3cm}}

\toprule
\textbf{\makecell[c]{Example \\ conversations }} & \textbf{Average politeness} & \textbf{Intuitive style attributes}\\
\midrule
\makecell[l]{\textsc{\textbf{\underline{Chipotle}}}\\\textbf{[Customer]} \$3 burritos and I’m nowhere near a Chipotle\\\textbf{[Agent]} Bummer. I'm so sorry. How far away is the clos-\\est location? –Becky} & $0.51$ & \makecell[l]{Friendly,\\Conversational,\\Not-impolite}\\
\midrule

 \makecell[l]{\textsc{\textbf{\underline{Comcast}}}\\\textbf{[Customer]} My internet is down and xfinity talkin about\\24-72 hours... y’all have the game messed up.\\\textbf{[Agent]} I understand this is a frustrating experience,\\please send a DM with your account information so I can\\look into this matter for you} & $0.77$ & \makecell[l]{Formal,\\Task-oriented}\\
\midrule

 \makecell[l]{\textsc{\textbf{\underline{VirginTrains}}}\\\textbf{[Customer]} See attached error message. I've tried leaving\\a voicemail several times in the past week.\\\textbf{[Agent]} Have you tried from another device?} & $0.50$ & \makecell[l]{Direct,\\To-the-point,\\Bot-like}\\

\bottomrule
\end{tabular}}
  
  \caption{Example of conversations of customer care agents from the TWCS dataset that show the limitation of style definitions using fixed attributes, here Politeness. Chipotle and VirginTrains customer care agents get roughly the same politeness score by an off-the-shelf politeness classifier \cite{danescu2013computational}, however, intuitively their style attributes are different as shown in the third column. 
  }
  \label{tab:off-the-shelf-example}
\end{table}

Our proposed \textit{few-shot conversation style transfer} task addresses several key challenges. Firstly, the interpretation of style attributes of the source/target dialogues is no longer required rather the style is defined solely through a few example dialogues. Secondly, it does not require parallel data in the form of source-to-target pairs, which is expensive and difficult to collect. Finally, conversation style transfer is performed with only a few example dialogues in the target style. In this paper, we show that transferring the conversation style in such a setting helps downstream applications such as chatbot personalization and domain adaptation for training.

\begin{figure*}[t!]
  \centering
  \includegraphics[width=0.8\textwidth]{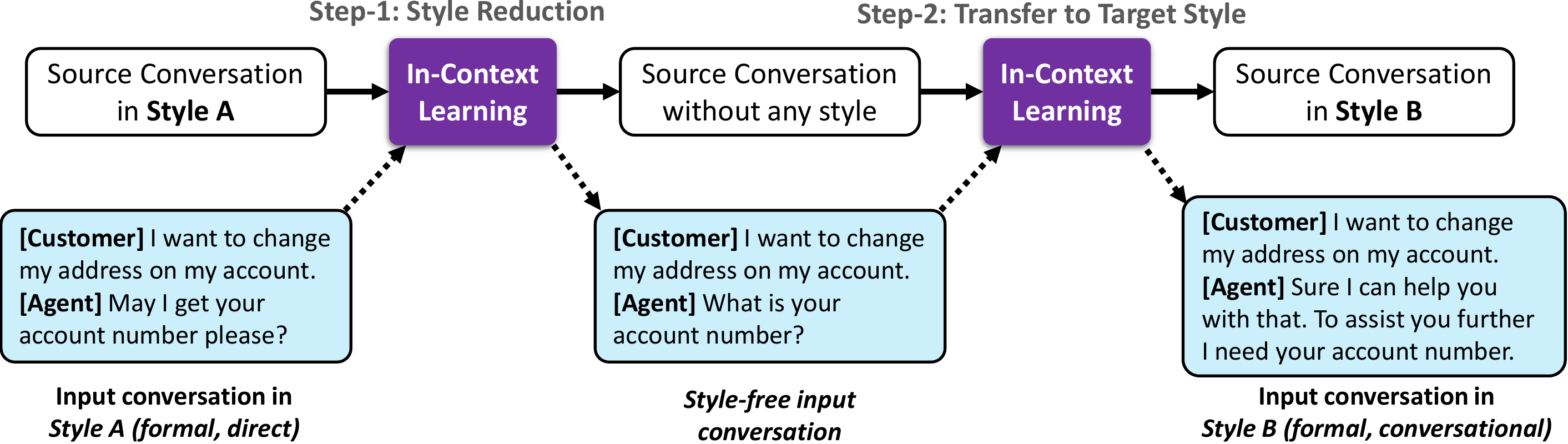}
  
  \caption{The proposed two-step in-context learning-based approach for conversation style transfer: \textit{(Step 1)}  The style in the source conversation is reduced and it is converted to a style-free conversation. \textit{(Step 2)} The style-free conversation is converted to the target style. Both conversion steps are learned in context.}
  
  \label{fig:modeling}
\end{figure*}

Tailored for the proposed few-shot learning problem, we propose a novel method based on in-context learning~\cite{brown2020language}. We propose to perform source-to-target style transfer with style-free dialogues as pivots. In this approach, we first prompt pre-trained large language models (LLMs) to perform style reduction on source dialogue, then use another set of prompts to rewrite the style-free dialogue to match the target style (Figure \ref{fig:modeling}).

To accurately and efficiently evaluate the quality of conversation style transfer using different models, we conduct 
human evaluation on style strength, appropriateness, and semantic correctness. The appropriateness assessment is unique to conversation style transfer, which evaluates whether the transferred utterances are out-of-context. Appropriateness is critical for Task-Oriented Dialogue (TOD) applications as inappropriate responses (as shown in Figure \ref{fig:problem-formulation}) can result in degraded user experience. As supplementary metrics, we report automatic scores on classifier-based style strength and semantic similarity. We observe that utterance-level style transfer without contextual information can achieve the highest style strength scores, however, results in low appropriateness and low semantic correctness. On the other hand, by including contextual information, although, with lower style strength, the transferred utterances are more appropriate and semantically correct.

Conversation style transfer can be applied in downstream tasks as a data augmentation or domain adaptation technique. We perform an extrinsic evaluation of style transfer in such a setting for intent classification task, where the training and test data for the task are from different style domains. We apply few-shot conversation style transfer on the training data to convert it to the test style before training. As a result, we observe improvement in intent classification F1 scores across three domains, demonstrating the usefulness of style transfer of conversations in such downstream applications.




\section{Problem Formulation: Few-Shot Conversation Style Transfer}\label{sec:problem-formulation}
In this section, we propose the novel task of conversation style transfer, based on a few non-parallel examples, that does not rely on style attribute definitions (an example is illustrated in Figure \ref{fig:problem-formulation}). Given a conversation in source style A and a few shot non-parallel conversations in target style B, the task is to transfer the style of the conversation in source style A to style B. We address the following limitations of the state-of-the-art models in this task.

    \paragraph{Few-shot availability of the target style examples:}Most of the existing works in style transfer assume that a large amount of text is available in the target style to train a model \citep{niu2018multi, wang2019harnessing, zhang2018style, briakou2021ola, madaan2020politeness, cheng2020contextual, reif2022recipe}. But this assumption may not hold in real-world settings.
    Hence, we limit target style data availability to a few dialogues. 
    
    \paragraph{Style transfer to arbitrary style:}Existing works explicitly define style attributes (e.g., politeness) and transfer a text with a \textit{known} style attribute to a style with another \textit{known} attribute, for example, impolite to polite \cite{madaan2020politeness}. However, the style of a conversation can be difficult to define with a fixed set of attributes as shown in Table \ref{tab:off-the-shelf-example}, and conversation style may be a combination of many attributes as conversations are dynamic. Hence, we study the problem of style transfer of conversations where the style attributes of the source and the target styles are \textit{not necessarily known}.
    
    
    \paragraph{Non-parallel 
    examples:}To train a model for transferring the style of a conversation from a source to a target style with a few examples, ideally, we want parallel conversations in the source and the target styles \cite{reif2022recipe,suzgun2022prompt}. However, parallel data is difficult to obtain and scale to many styles (including out-of-domain styles) due to challenges in determining conversational style attributes and stylizing conversations.
    Hence, we assume access to a few examples in the source and the target styles that are \textit{not parallel}.
    
    \paragraph{Evaluation criteria:}A successful conversation style transfer model is expected to produce dialogues matching the target style, while preserving the original semantics and appropriateness of the turns. So in this paper, we evaluate our models on the following metrics.
    \begin{itemize}
        \item \textbf{Style strength:} Following previous studies \citep{reif2022recipe,han-etal-2022-meet} we evaluate the target style strength of utterances produced by a style transfer model. The style strength scores are higher if the transferred utterances match the target style. 
        
        \item \textbf{Semantic correctness:} In the context of TODs, we define semantic correctness as the preservation of intents in style-transferred conversations.
        
        \item \textbf{Appropriateness of response:} Appropriateness of response can be defined as the coherence of a 
        response given the previous turns in a conversation. This is required in TODs to prevent the style-transferred utterances in a dialogue from being out-of-context.
    \end{itemize}
    Positive and negative examples of these metrics are shown in Table \ref{tab:evaluation-metrics}.
    

\begin{table*}
\centering
\resizebox{2\columnwidth}{!}{%
\begin{tabular}
{>{\centering\arraybackslash}m{5.5cm}|>{\centering\arraybackslash}m{5.5cm}>{\centering\arraybackslash}m{5.5cm}>{\centering\arraybackslash}m{5.5cm}}

\toprule

\multicolumn{1}{c}{\textbf{Chatbot Style (Crisp/Direct)}} & \multicolumn{3}{c}{\textbf{Human Style (Conversational)}}\\
\cmidrule(lr){1-1}\cmidrule(lr){2-4}
\textbf{Source Conversation} & \textbf{Style Transferred v1} & \textbf{Style Transferred v2} & \textbf{Style Transferred v3}\\
\cmidrule(lr){1-1}\cmidrule(lr){2-2}\cmidrule(lr){3-3}\cmidrule(lr){4-4}
\makecell[l]{\textbf{[Customer]} I had an accident and I\\want to file an auto insurance claim.\\\textbf{[Agent]} What is your insurance\\number?}
& \makecell[l]{\textbf{[Customer]} I had an accident and I\\want to file an auto insurance claim.\\\textbf{[Agent]} I am sorry to hear that. Can\\I get your insurance number?}
& \makecell[l]{\textbf{[Customer]} I had an accident and I\\want to file an auto insurance claim.\\\textbf{[Agent]} \textcolor{red}{I am happy to hear that}.\\Can I get your insurance number?}
& \makecell[l]{\textbf{[Customer]} I had an accident and I\\want to file an auto insurance claim.\\\textbf{[Agent]} Can I get \textcolor{red}{my} insurance\\number?}\\
\cmidrule(lr){1-1}\cmidrule(lr){2-2}\cmidrule(lr){3-3}\cmidrule(lr){4-4}

\makecell[l]{Human Style Strength: Low\\Appropriate? - Yes}
& \makecell[l]{Human Style Strength: High\\Appropriate? - Yes\\Semantically Correct? - Yes}
& \makecell[l]{Human Style Strength: High\\Appropriate? - No\\Semantically Correct? - Yes}
& \makecell[l]{Human Style Strength: Low\\Appropriate? - Yes\\Semantically Correct? - No}\\

\bottomrule
\end{tabular}}
\caption{Example of style transfer evaluation metrics - \textit{style strength, appropriateness}, and \textit{semantic correctness}, by comparing three style transferred versions of the same agent utterance.
Inappropriate and semantically incorrect segments of the generated utterances are marked in red. 
}
\label{tab:evaluation-metrics}
\end{table*}
\section{In-Context Learning for Conversation Style Transfer}\label{sec:modeling}
In this section, we propose a novel in-context learning based method using large language models (LLMs) for few-shot conversation style transfer. The method is illustrated in Figure \ref{fig:modeling}.


\subsection{In-context learning with non-parallel examples in source and target styles}\label{subsec:prompt-non-parallel}


To tackle the problem of the unavailability of parallel conversations in source and target styles (as described in Section \ref{sec:problem-formulation}), in this paper, we propose a cheaper alternative solution, which prompts the language models with dialogues in one style and their style-free versions. 
Previous work by \citet{madaan2020politeness} showed the effectiveness of style transfer after reducing the source text to a style-free format and then converting the style-free format to the target style (although they relied on large amount of training data for the purpose). Inspired from these intuitions we break down the task of style transfer in the following two steps.
 
 \begin{enumerate}[itemsep=2pt]
     \item \textbf{Style Reduction:} In this step, we use an in-context learning method using LLMs to reduce the source conversation to a style-free form.
     As a result, we need parallel examples only in the form ($C_A$, $C^\prime$) for prompting LLMs, where $C_A$ is a conversation in source style A and $C^\prime$ is the style free form of $C_A$. 
     
     \item \textbf{Transfer to the Target Style: 
     } In this step, we use another in-context learning step where we convert the style-free input conversation to the target style. This step also requires parallel examples only in the form ($C^\prime$, $C_B$), where $C_B$ is a conversation in target style B and $C^\prime$ is the style free form of $C_B$.
     
 \end{enumerate}
 We use human supervision to construct the parallel ($C_{A/B}$, $C^\prime$) examples. Note that, it is easier for humans to rewrite a conversation in a style-free format as it omits the requirement of having knowledge about the target style. Prompt structures and examples for in-context learning for the above two steps can be found in Appendix \ref{appendix:prompting}.
 
 \subsection{Dynamic Prompt Selection}\label{subsection:dynamic-prompt-selection} 
 Conversations are dynamic and the style of a response depends on the situation as shown in Figure \ref{fig:problem-formulation}. Hence, the same set of few-shot examples may not work best as training examples for all test conversations as the situations and respective styles may be different (e.g., different styles are expected when responding to someone distressed vs happy). To resolve this, we propose a dynamic prompt selection technique \citep{reif2022recipe,han-etal-2022-meet} for style transfer where semantically similar examples to a test conversation are retrieved and used as few-shot training examples in the prompt. We first concatenate all utterances of a participant in a conversation sequentially. Then we use a sentence transformer \cite{reimers2019sentence} designed for semantic search to encode the concatenated utterances to get a semantic meaning-based embedding. For each test conversation, we measure the cosine similarity between the embedding of the test conversation and all of the available few-shot training conversations. We select the top-k semantically similar few-shot examples for the test conversation during prompting. The more semantically similar conversation appears later in the prompt to place it closer to the test conversation. The effectiveness of this approach is examined by comparing it with random prompt example selection method in Section \ref{sec:experiment}.

 \subsection{Baseline: Utterance level style transfer} Existing works study style transfer at the utterance level with in-context learning \citep{reif2022recipe,suzgun2022prompt}, hence, we use utterance-level style transfer as a baseline. We transfer the style of the utterances of one party in a dialogue utterance by utterance using the same procedure described above. For dynamic prompt selection, we measure semantic similarity between single utterances instead of concatenating all utterances of a participant in a dialogue. As existing models are either applicable to utterance level only \citep{riley2021textsettr} or require a lot of training data \citep{madaan2020politeness} for style transfer, they are not applicable in conversation style transfer in a few-shot setting. 

\section{Experiments}\label{sec:experiment}
In this section, we present the evaluation setup and the results of the proposed approaches on style transfer quality including style strength, appropriateness, and semantic correctness. Then, we show the evaluation results of applying the approach on a downstream task, namely intent classification.

\subsection{Setup} 

\paragraph{Dataset:}Given that our focus is on TODs, we extract conversations from the following two TOD datasets for studying style transfer. 

\begin{itemize}
    \item \textbf{TWCS dataset} \cite{stuart_axelbrooke_2017}: Contains real-life human customer care agent dialogues with customers of different companies.
    \item \textbf{Cross-domain conversational data from DSTC11 intent induction track}\footnote{\url{https://github.com/amazon-research/dstc11-track2-intent-induction}}: Contains human-to-human (human agents) and human-to-bot (bot agents) dialogues.
\end{itemize}
To study style transfer, we select human agents dialogues (addressed as $H_1$) and bot agents dialogues (addressed as $B$) from DSTC11. Then we select Chipotle customer care agent dialogues from TWCS as another human style (addressed as $H_2$).


\begin{table}[t!]
\centering
\resizebox{0.8\columnwidth}{!}{%
\begin{tabular}
{>{\arraybackslash}m{0.7cm}>{\centering\arraybackslash}m{2.2cm}>{\centering\arraybackslash}m{2.2cm}>{\centering\arraybackslash}m{1.8cm}}

    \toprule
    \textbf{Styles} & \textbf{Avg. agent turns / conv.} & \textbf{Avg. words / agent turn (Crispness)} & \textbf{Vocabulary size (Diversity)}\\
    \cmidrule(lr){1-4}
    $H_1$   & $35.84$ $(\pm9.6)$    & $11.62$ $(\pm8.9)$    & $6529$\\
    $B$     & $5.23$ $(\pm3.3)$     & $6.55$ $(\pm1.8)$     & $142$\\
    $H_2$   & $2.64$ $(\pm0.8)$     & $11.55$ $(\pm6.3)$    & $1698$\\
    \bottomrule
\end{tabular}}

\caption{Quantitative differences among styles $H_1$, $H_2$, and $B$. Human agents ($H_1$, $H_2$) are more conversational and use diverse words compared to bots ($B$). The bot style is very crisp and to-the-point. Apart from these properties, in human style, $H_2$ agents sign their names at the end of the response $98\%$ of the time.}
\label{tab:quantitative-style}
\vspace{1em}
\centering
\resizebox{1\columnwidth}{!}{%
\begin{tabular}
{>{\arraybackslash}m{0.9cm}>{\arraybackslash}m{12cm}}

    \toprule
    \textbf{Styles} & \textbf{High PMI style indicator lemmas}\\
    \cmidrule(lr){1-2}
    $H_1$   &  mister, alright, sorry, kindly, bye, mhm, uh, um, worry, huh, morning, pleasure, sir, goodbye, yes, fine, ok, afternoon, great, yeah, perfect, oh, sure, thank, glad\\
    \cmidrule(lr){1-2}
    $B$     & please, welcome, hello\\
    \cmidrule(lr){1-2}
    $H_2$   & cool, inconvenience, apology, wow, totally, fan, asap, frustrating, unfortunately, hey, disappointing, awesome, troubling, guy, shoot, gonna, ah, gotcha, friend, love, appreciate, bummer, happy, definitely, hope\\
    \bottomrule
\end{tabular}}

\caption{High PMI style indicator lemmas for each style domain (details on PMI calculation can be found in Appendix \ref{appendix:pmi-lemma-procedure}). We can observe that chatbots ($B$) are crisp and do not use many non-topic-specific words. Mostly formal words are used in human style $H_1$, and many informal and friendly words (e.g., bummer) are used in human style $H_2$. Example conversations of each style can be found in Appendix \ref{appendix:example-conversations}.}

\label{tab:style-lexicon}
\end{table}

We observe that the three styles, $H_1$, $H_2$, and $B$ are holistically different. Some observed properties of the human styles ($H_1$, $H_2$) are that they are engaging, conversational, and use diverse vocabulary (Table \ref{tab:quantitative-style}). Being conversational and engaging, humans can be formal or casual and may use different structures for their responses. For example, human style $H_1$ is formal (uses formal words such as `mister') while the other human style $H_2$ is casual and friendly (uses millennial phrases such as `cool', `asap'). Additionally, in human style $H_2$, human agents sign their names at the end of each response, implying a structural stylistic property of this human style. Some observed properties of the bot style are crispness and to-the-point while not being informal. These observed properties are summarized with quantitative and qualitative analyses in Tables \ref{tab:quantitative-style} and \ref{tab:style-lexicon}, and example conversations in these styles are presented in Appendix \ref{appendix:example-conversations}. This analysis supports our claim that conversation styles are holistic and difficult to characterize using a fixed set of attributes. 

We study style transfer with the three complex styles stated above where we are able to evaluate the style transfer performance on drastically different style pairs (e.g., human style $H_1$/$H_2$ to/from bot style $B$), as well as pairs with nuanced differences (e.g., human style $H_1$ to human style $H_2$). The style directions we study in this paper and respective dataset statistics are shown in Table \ref{tab:dataset-summary}.


\paragraph{In-context Learning:} We conduct in-context learning experiments with two decoder-only LLMs - GPT NeoX \cite{black2022gpt} (20B parameters) and Bloom\footnote{\url{https://huggingface.co/bigscience/bloom}} (176B parameters). Details of the LLMs can be found in Appendix \ref{appendix:llm-hyperparameters}.

\begin{table}[t!]
\centering
\resizebox{1\columnwidth}{!}{%
\begin{tabular}
{>{\arraybackslash}m{1.6cm}>{\centering\arraybackslash}m{1.4cm}|>{\centering\arraybackslash}m{1cm}>{\centering\arraybackslash}m{1.8cm}|>{\centering\arraybackslash}m{1cm}>{\centering\arraybackslash}m{1.8cm}}

\toprule
    \multicolumn{1}{c}{} & \multicolumn{1}{c}{} & \multicolumn{2}{c}{\textbf{Validation Set}} &  \multicolumn{2}{c}{\textbf{Test Set}} \\
    \cmidrule(lr){3-4}\cmidrule(lr){5-6}
    \textbf{\makecell[c]{Style\\Directions}} & \textbf{\# conversations} & \textbf{\# segments} & \textbf{\# agent utterances} & \textbf{\# segments} & \textbf{\# agent utterances}\\
   \cmidrule(lr){1-1} \cmidrule(lr){2-2} \cmidrule(lr){3-4}\cmidrule(lr){5-6}
    $H_1 \rightarrow B$ & $25$ & $201$ & $497$ & $65$ & $164$\\
    $H_1 \rightarrow H_2$ & $25$ & $201$ & $495$ & $65$ & $166$\\
    $B \rightarrow H_1$ & $25$ & $37$ & $90$ & $65$ & $152$\\
    $B \rightarrow H_2$ & $25$ & $37$ & $90$ & $65$ & $152$\\
    \bottomrule
\end{tabular}}
\caption{Validation and test data statistics. Long conversations are divided into small segments consisting of 4-5 turns. We cover four style transfer directions to/from two human styles ($H_1$, $H_2$) and bot style ($B$).}

\label{tab:dataset-summary}
\end{table}

\begin{table*}[ht!]
\centering
\resizebox{1.7\columnwidth}{!}{%
\begin{tabular}{lcccccc}
\toprule
    \multicolumn{1}{c}{} & \multicolumn{3}{c}{\textbf{\textsc{GPT-NeoX (20B)}}} &  \multicolumn{3}{c}{\textbf{\textsc{Bigscience-Bloom (176B)}}} \\
    \cmidrule(lr){2-4}\cmidrule(lr){5-7}
    \textbf{\makecell[c]{Style\\Directions}} & \textbf{Appropriateness} & \textbf{Style Strength} & \textbf{Semantic Correct.} & \textbf{Appropriateness} & \textbf{Style Strength} & \textbf{Semantic Correct.}\\
   \cmidrule(lr){1-1}\cmidrule(lr){2-4}\cmidrule(lr){5-7}
    $H_1 \rightarrow B$ & 0.98 (0.06) & 0.88 (0.26) & 0.80 & 0.96 (0.15) & 0.82 (0.33) & 0.78\\
    $H_1 \rightarrow H_2$ & 0.97 (0.06) & 0.69 (0.31) & 0.87 & 0.96 (0.08) & 0.81 (0.23) & 0.76\\
    $B \rightarrow H_1$ & 1 (0.02) & 0.86 (0.12) & 0.95 & 0.98 (0.08) & 0.75 (0.27) & 0.87\\
    $B \rightarrow H_2$ & 0.97 (0.05) & 0.90 (0.08) & 0.99 & 0.97 (0.14) & 0.91 (0.08) & 0.89\\
    \bottomrule
\end{tabular}}
\caption{Inter-annotator agreement scores for the three human evaluation tasks. Standard deviations over all data points are shown in brackets for the style strength and appropriateness evaluation tasks. The detailed procedure for calculating the agreement scores can be found in Appendix \ref{appendix:human-evaluation}.}

\label{tab:inter-annotator-agreement}
\end{table*}

\paragraph{Prompt Settings:}We tune two hyperparameters in the prompt: (1) the number of contextual turns from the dialogue history, (2) the number of examples in the prompt. For the number of contextual turns, we experiment with short segments (2 turns) and long segments (4-5 turns). For the number of examples, we select the hyperparameter based on the validation set (Table \ref{tab:dataset-summary})\footnote{We experiment with 5, 10, 20 examples in the prompt for utterance level style transfer and short segments, and 4, 8 examples for long segments on validation set. The best hyperparameters were 10, 10, and 8 for utterance-level, short segment, and long segment, respectively (Tab. \ref{tab:appendix-ablation}, Appx. \ref{appendix:ablation-study}).}.  Note that when increasing the number of turns further, the maximum context length of LLMs is reached quickly, therefore, we leave in-context learning with full dialogue context as a future work. In Appendix \ref{appendix:prompting}, we show example prompts for baseline (utterance-level), short-segment, and long-segment.

%

\paragraph{Construction of Few-Shot Examples:} 
As mentioned in Section~\ref{subsec:prompt-non-parallel}, we construct a few (styled, style-free) conversation pairs for each style domain using human supervision. Comparing the creation of true parallel data between source and target styles, such an approach is easy to execute for humans and results in reusable examples. 
Humans were asked to reduce the style of the whole conversation. It took approximately 5 minutes for a human to rewrite a 10-12 turns conversation to a style-free form. As the style reduction is a cheap one-time effort in our work, we leave automatic style reduction as a future work. The human annotation method, statistics, and examples can be found in Appendix \ref{appendix:creation-of-few-shot-examples}.

\begin{table}[t!]
\centering

\resizebox{0.8\columnwidth}{!}{%
\begin{tabular}
{>{\arraybackslash}m{1.5cm}>{\arraybackslash}m{1.6cm}>{\centering\arraybackslash}m{1.5cm}|>{\centering\arraybackslash}m{1.5cm}>{\centering\arraybackslash}m{1.5cm}}
\toprule
    \multicolumn{1}{c}{} & \multicolumn{1}{c}{} &  \multicolumn{3}{c}{\textbf{\makecell[c]{Target style strength}}} \\ 
    \cmidrule(lr){3-5}
    \multicolumn{1}{c}{} & \multicolumn{1}{c}{} & \multicolumn{1}{c}{\textbf{Before} } &  \multicolumn{2}{c}{\textbf{\makecell[c]{After} / Prompt selection}}\\
    \textbf{Models (\# shots)} & \textbf{Style directions} & --  & \textbf{Random} & \textbf{Dynamic}\\
    \cmidrule(lr){1-5}
    \multirow{5}{*}{\textbf{\makecell[l]{Utterance\\level style\\transfer\\(10 shots)}}}
    & $H_1 \rightarrow B$    & $0.010$ & $0.077$ & $0.150$\\
    & $H_1 \rightarrow H_2$   & $0.112$ & $0.182$ & $0.215$\\
    & $B \rightarrow H_1$   & $0.001$ & $0.411$ & $0.556$\\
    & $B \rightarrow H_2$   & $0$     & $0.337$ & $0.671$\\
    \cmidrule(lr){2-5}
    & Average           & $0.031$ & $0.252$ & $\bm{0.398}$\\
    \cmidrule(lr){1-5}
    
    \multirow{5}{*}{\textbf{\makecell[l]{2-turns\\conv. level\\style tran.\\(10 shots)}}}
    & $H_1 \rightarrow B$    & $0.010$ & $0.045$ & $0.119$\\
    & $H_1 \rightarrow H_2$   & $0.112$ & $0.165$ & $0.199$\\
    & $B \rightarrow H_1$   & $0.001$ & $0.101$ & $0.399$\\
    & $B \rightarrow H_2$   & $0$     & $0.062$ & $0.113$\\
    \cmidrule(lr){2-5}
    & Average           & $0.031$ & $0.093$ & $\bm{0.208}$\\
    \cmidrule(lr){1-5}
    
    \multirow{5}{*}{\textbf{\makecell[l]{4/5-turns\\conv. level\\style tran.\\(8 shots)}}}
    & $H_1 \rightarrow B$    & $0.010$ & $0.100$ & $0.160$\\
    & $H_1 \rightarrow H_2$   & $0.112$ & $0.165$ & $0.173$\\
    & $B \rightarrow H_1$   & $0.001$ & $0.291$ & $0.420$\\
    & $B \rightarrow H_2$   & $0$     & $0.058$ & $0.110$\\
    \cmidrule(lr){2-5}
    & Average           & $0.031$ & $0.154$ & $\bm{0.216}$\\
    \bottomrule
\end{tabular}}
\caption{Comparison between dynamic and random prompt selection on target style strength across utterance and conversation level style transfers.}

\label{tab:prompt-selection-ablation}
\end{table}

\begin{table*}[t!]
\centering

\resizebox{2\columnwidth}{!}{%
\begin{tabular}
{>{\centering\arraybackslash}m{0.5cm}|>{\arraybackslash}m{1.6cm}>{\centering\arraybackslash}m{1.6cm}>{\centering\arraybackslash}m{2.6cm}>{\centering\arraybackslash}m{2.5cm}>{\centering\arraybackslash}m{2.5cm}|>{\centering\arraybackslash}m{1.6cm}>{\centering\arraybackslash}m{2.6cm}>{\centering\arraybackslash}m{2.5cm}>{\centering\arraybackslash}m{2.5cm}}
\toprule
    \multicolumn{2}{c}{} & \multicolumn{4}{c}{\textbf{GPT-NeoX (20B)}} &  \multicolumn{4}{c}{\textbf{Bloom (176B)}} \\
    \cmidrule(lr){3-6}\cmidrule(lr){7-10}
   & \textbf{Style} & \textbf{Original} & \textbf{Utterance Level} & \multicolumn{2}{c}{\textbf{{Conversation Level}}} & \textbf{Original} & \textbf{Utterance Level} & \multicolumn{2}{c}{\textbf{{Conversation Level }}}\\
   \cmidrule(lr){5-6}\cmidrule(lr){9-10}
   & \textbf{Directions} & \textbf{Utterances} & \textbf{1 turn} & \textbf{2 turns} & \textbf{4/5 turns} & \textbf{Utterances} & \textbf{1 turn} & \textbf{2 turns} & \textbf{4/5 turns}\\
   \cmidrule(lr){1-2}\cmidrule(lr){3-6}\cmidrule(lr){7-10}
    
    \multirow{4}{*}{\textbf{\rotatebox[origin=c]{90}{Style Strength}}}
    & $H_1 \rightarrow B$ & $0.392$ & $0.864$ & $0.714$ & $0.561$ & $0.435$ & $0.876$ & $0.719$ & $0.720$\\
      & $H_1 \rightarrow H_2$ & $0.15$ & $0.854$ & $0.855$ & $0.838$ & $0.125$ & $0.895$ & $0.924$ & $0.538$\\
      & $B \rightarrow H_1$ & $0.574$ & $0.851$ & $0.846$ & $0.690$ & $0.378$ & $0.692$ & $0.622$ & $0.856$\\
      & $B \rightarrow H_2$ & $0.043$ & $0.989$ & $0.805$ & $0.690$ & $0.024$ & $0.958$ & $0.897$ & $0.484$\\
    \cmidrule(lr){2-6}\cmidrule(lr){7-10} 
    & Average & $0.290$ & $\bm{0.890}$ & $0.805$ & $0.695$ & $0.241$ & $\bm{0.855}$ & $0.791$ & $0.650$\\
    \cmidrule(lr){1-10}
    
    
    \multirow{4}{*}{\textbf{\rotatebox[origin=c]{90}{Appropriate.}}}
    & $H_1 \rightarrow B$ &  $0.997$ & $0.943$ & $0.971$ & $0.979$ & $0.991$ & $0.968$ & $0.974$ & $0.966$\\
      & $H_1 \rightarrow H_2$ & $0.980$ & $0.798$ & $0.985$ & $0.977$ & $0.997$ & $0.917$ & $0.972$ & $0.974$ \\
      & $B \rightarrow H_1$ & $0.997$ & $1.0$ & $0.997$ & $0.987$ & $0.995$ & $0.995$ & $0.980$ & $0.968$\\
      & $B \rightarrow H_2$ & $0.990$ & $0.481$ & $1.00$ & $0.978$ & $0.995$ & $0.923$ & $0.957$ & $0.976$\\
    \cmidrule(lr){2-6}\cmidrule(lr){7-10} 
    & Average & $0.991$ & $0.806$ & $\bm{0.988}$ & $0.980$ & $0.995$ & $0.951$ & $\bm{0.971}$ & $\bm{0.971}$\\
    \cmidrule(lr){1-10}

    
    \multirow{5}{*}{\textbf{\rotatebox[origin=c]{90}{Semantic Correct.}}}
    & & & \textbf{yes-partial-no} & \textbf{yes-partial-no} & \textbf{yes-partial-no} & & \textbf{yes-partial-no} & \textbf{yes-partial-no} & \textbf{yes-partial-no}\\
    \cmidrule(lr){4-6}\cmidrule(lr){8-10}
    & $H_1 \rightarrow B$ &  & $0.89$-$0.02$-$0.09$ & $0.94$-$0.01$-$0.05$ & $0.92$-$0.03$-$0.05$ &  & $0.95$-$0$-$0.05$ & $0.97$-$0$-$0.03$ & $0.77$-$0.03$-$0.20$\\
    & $H_1 \rightarrow H_2$ &  & $0.92$-$0.01$-$0.07$ & $0.96$-$0.01$-$0.03$ & $0.94$-$0.02$-$0.04$ &  & $0.89$-$0.01$-$0.10$ & $0.96$-$0$-$0.04$ & $0.84$-$0.01$-$0.15$\\
    & $B \rightarrow H_1$ &  & $1.00$-$0$-$0$ & $0.98$-$0$-$0.02$ & $0.96$-$0.02$-$0.02$ &  & $1.00$-$0$-$0$ & $0.97$-$0$-$0.03$ & $0.86$-$0$-$0.14$\\
    & $B \rightarrow H_2$ &  & $0.99$-$0$-$0.01$ & $1.00$-$0$-$0$ & $1.00$-$0$-$0$ &  & $1.00$-$0$-$0$ & $0.99$-$0$-$0.01$ & $0.88$-$0$-$0.12$\\
    \cmidrule(lr){2-6}\cmidrule(lr){7-10} 
    & Average & & $0.95$-$0.01$-$0.04$ & $\bm{0.97}$-$\bm{0}$-$\bm{0.03}$ & $0.96$-$0.01$-$0.03$ & & $0.96$-$0$-$0.04$ & $\bm{0.97}$-$\bm{0}$-$\bm{0.03}$ & $0.84$-$0.01$-$0.15$\\
    \bottomrule
\end{tabular}}
\caption{Human evaluation results for utterance level (baseline) and conversation level style transfer with GPT-NeoX and Bloom LLMs using our model. The best average scores over all style dimensions are marked in bold. Utterance level style transfer achieves higher style strength but conversation level style transfers yield more appropriate and semantically correct responses. Statistics with standard deviations can be found in Appendix \ref{appendix:style-transfer-eval-results-with-sd}.}

\label{tab:human-evaluation-results}
\end{table*}

\begin{table*}[h]
\centering
\resizebox{1.9\columnwidth}{!}{%
\begin{tabular}
{>{\centering\arraybackslash}m{0.5cm}|>{\raggedleft\arraybackslash}m{3.2cm}>{\raggedright\arraybackslash}m{13cm}>{\raggedright\arraybackslash}m{1.5cm}>{\centering\arraybackslash}m{1.5cm}>{\centering\arraybackslash}m{1.5cm}>{\centering\arraybackslash}m{1.5cm}}

\toprule
    \multicolumn{1}{c}{} & \multicolumn{1}{c}{} & \multicolumn{1}{c}{} & & \multicolumn{3}{c}{\textbf{Human Evaluation Scores}} \\
    \cmidrule(lr){5-7}
    & & \textbf{Example Conversations} & \textbf{Models} & \makecell[c]{Style\\Strength} & \makecell[c]{Appropr-\\iateness} & \makecell[c]{Semantic\\Correct.} \\ 
    \cmidrule(lr){1-7}
    
    \multirow{5}{*}{\textbf{\rotatebox[origin=c]{90}{Example-1}}} &
    \textbf{Previous Turn:} & \textbf{[Customer]} I don't think so. Oh! Yes, when will I get new cards for my son? & & & &\\
    & \textbf{Original Turn ($H_1$):} & \textbf{[Agent]} let me check on that for your & & 0.33 & 1 & -\\
    \cmidrule(lr){3-3}
    & \multirow{3}{*}{\textbf{Transferred ($H_2$):}} 
    & \textbf{[Agent]} I'm sorry to hear that. Can you please let me check on that for you? –AC & 1 turn & $1$ & $0$ & Yes\\
    & & \textbf{[Agent]} Oh, I will check on that for you. –AC & 2 turns  & $0.72$ & $1$ & Yes\\
    & & \textbf{[Agent]} I will check on that for you. –Gabe & 4/5 turns  & $0.61$ & $1$ & Yes\\
    \cmidrule(lr){1-7}
    
    \multirow{5}{*}{\textbf{\rotatebox[origin=c]{90}{Example-2}}} &
    \textbf{Previous Turn:} & \textbf{[Customer]} The purpose is to check if I have cancer. & & & &\\
    & \textbf{Original Turn ($B$):} & \textbf{[Agent]} May I have your email address?  & & 0.33 & 1 & -\\
    \cmidrule(lr){3-3}
    & \multirow{3}{*}{\textbf{Transferred ($H_1$):}} 
    & \textbf{[Agent]} What's your email address? & 1 turn & $0.33$ & $1$ & Yes\\
    & & \textbf{[Agent]} Hi, my name is Jane, and I would like to check if you have cancer. & 2 turns  & $0.50$ & $0.33$ & No\\
    & & \textbf{[Agent]} Perfect. And what is your email address? & 4/5 turns  & $1$ & $0.33$ & Yes\\
    \cmidrule(lr){1-7}
    
    \multirow{5}{*}{\textbf{\rotatebox[origin=c]{90}{Example-3}}} &
    \textbf{Previous Turn:} & \textbf{[Customer]}  Could you repeat that? Sorry. & & & &\\
    & \textbf{Original Turn ($H_1$):} & \textbf{[Agent]} No problem! three two three three seven eight four two one one & & $0.33$ & $1$ & -\\
    \cmidrule(lr){3-3}
    & \multirow{3}{*}{\textbf{Transferred ($H_2$):}} 
    & \textbf{[Agent]} Yes! -AC & 1 turn & $0.67$ & $0.67$ & No\\
    & & \textbf{[Agent]} No problem! Three two three three seven eight four two one one. -AC & 2 turns & $1$ & $1$ & Yes\\
    & & \textbf{[Agent]}  No problem. Three2384211. –James & 4/5 turns  & $1$ & $1$ & Yes\\
    \bottomrule
\end{tabular}}

\caption{Qualitative error analysis on hand-picked examples. Human evaluation scores associated with the output of various versions of the model (using Bigscience-Bloom as the LLM) are shown in the three rightmost columns. A score of 1 means the most appropriate or most target style strength.}
\label{tab:error-analysis}
\end{table*}

\paragraph{Automatic Evaluation:} To measure the strength of the target style automatically, we train RoBERTa \cite{liu2019roberta} based binary text classifiers to classify between the source style and the target style. Training data for these classifiers were obtained from conversational data with both styles. The validation accuracy of the classifiers to differentiate between styles ($H_{1}$, $B$), ($H_{1}$, $H_{2}$), and ($H_{2}$, $B$) were $99.89\%$, $93.3\%$ and $100\%$, respectively. The details on these classifiers can be found in Appendix \ref{appendix:style-discriminator}. We treat the confidence scores of the classifiers as the style strength scores. For semantic similarity we measure the cosine distance between SBERT embeddings \cite{reimers2019sentence} of a source utterance and the corresponding style transferred utterance. For the evaluation of appropriateness, we rely only on human evaluation as it is difficult to get an automatic method to measure appropriateness.

\paragraph{Human Evaluation:}To obtain a direct assessment of the style transfer quality of different models efficiently, we perform a ranking-based human evaluation on style strength and appropriateness. To evaluate style strength, we present human evaluators with utterances in the target style to train them on the properties of the target style. Then we present them with a source utterance and the style transferred versions of it by our proposed models and the baseline. The model names are kept hidden from them and the order of the utterances are shuffled. Then we ask the evaluators to rank all versions of the same utterance in a descending order based on the style similarity with the reference utterances. To evaluate appropriateness, we present human evaluators with a source agent utterance and all versions of the style transferred utterances along with the immediate previous customer turn as context. Then we ask the evaluators to rank them based on the appropriateness of the agent response. To evaluate semantic correctness, we present human evaluators with a source utterance and the corresponding style transferred utterances. We ask them for each style transferred version if it is semantically similar, partially similar, or dissimilar to the source utterance. Each data point is evaluated by three human evaluators who are professional data linguists. We do not include data points where all the models generated exactly the same response. The inter-annotator agreement scores for the three human evaluation tasks are presented in Table \ref{tab:inter-annotator-agreement}. We convert the rankings of the evaluators to a scale of 1 where a higher score means a higher rank (i.e., more appropriate or more similar in style). To aggregate scores we average ranking scores by three evaluators. The pairwise comparison statistics among the models can be found in Appendix \ref{appendix:pairwise-comparison}. For semantic correctness, we select the label by taking majority voting. Details on human evaluation data statistics, evaluation interfaces, inter-annotator agreement scores calculation, and rank-scaling can be found in Appendix \ref{appendix:human-evaluation}.

\paragraph{Ablation Study:} We compare dynamic prompt selection with random prompt selection as described in Section \ref{subsection:dynamic-prompt-selection}. With the ablation on automatic style strength metric using GPT-NeoX, we find that dynamic prompt selection outperforms the random prompt selection method by a large margin as shown in Table \ref{tab:prompt-selection-ablation}.




\subsection{Results}
We show human evaluation results on utterance-level and conversation-level style transfer in Table~\ref{tab:human-evaluation-results}. Models were run on test data (Table \ref{tab:dataset-summary}) using the best hyper-parameters and prompt selection method obtained in the ablation step. We first observe that the highest style strength rank score is achieved when performing utterance-level style transfer, however, this results in a lower appropriateness score. This observation shows that performing conversation style transfer without the dialogue context has a significant risk of resulting in inappropriate agent utterances (i.e., utterances do not fit in the context). We can also observe in Table \ref{tab:human-evaluation-results} that the smaller LLM GPT-NeoX suffers more from the problem of generation of inappropriate responses compared to the larger LLM Bloom. Next, we observe that if we increase context (4/5 turns) in the conversation style transfer, the style strength decreases but appropriateness is preserved. Interestingly, for the larger LLM Bloom, the semantic similarity decreases with the increase of context. We found out that sometimes Bloom generates new agent utterances different from the source utterances or swaps the agent utterance with the customer utterance when performing 4-5 turns conversation-level style transfer (examples are shown in Appendix \ref{appendix:error-analysis-bigscience-bloom}). Hence, resulting in semantically dissimilar utterances. 

Therefore, we conclude that the LLMs are still not successful in conditioning on a larger context when performing style transfer, hence, a limited context consisting of 2 utterances is the optimal setting for style transfer in our study. Automatic evaluation results on the test set resulted in the same pattern (shown in Appendix \ref{appendix:style-transfer-eval-results-with-sd}). Examples of style-transferred conversations in all style directions by various versions of our model are shown in Appendix \ref{appendix:qualitative-examples} and the effects of style transfer on the observed style properties in Table \ref{tab:quantitative-style} are discussed in Appendix \ref{appendix:effect-on-observed-properties}. We present examples of errors by various versions of the models in Table \ref{tab:error-analysis}.

\subsection{Evaluation on Downstream Task}\label{sec:downstream-application}
Downstream applications of conversation style transfer are understudied. In this paper, we apply conversation style transfer to intent classification. 
We evaluate the setting where we have abundant of training data in one style and the test data is in a different style. We test our approach on three domains in the DSTC11 intent induction dataset: insurance, banking, and finance. Here, the training data is in human-to-human (h2h) style and the test data is in human-to-bot (h2b) style. We transfer the training data from h2h style to h2b style before training a RoBERTa-based intent classifier.

We run an ablation (using data from banking and finance domains) with utterance-level style transfer and short-conversation-level style transfer using GPT-NeoX and observe that training data transferred to h2b style using utterance-level style transfer results in higher intent classification F1 scores. We conjecture the reason is that utterance-level style transfer has the strongest style strength score, benefiting the application of domain adaptation. We report results with this method on all three domains in Table \ref{tab:ic-results-f1-score}. The intent classification results show statistically significant improvement in insurance and banking, and non-significant improvement in finance, compared to the baseline where the training data has h2h style.
Data statistics, experimental details, and ablation studies can be found in Appendix \ref{appendix:downstream-application}. 



\begin{table}[t]
\centering
\resizebox{1\columnwidth}{!}{%
\begin{tabular}
{>{\arraybackslash}m{3.8cm}>{\centering\arraybackslash}m{1.8cm}>{\centering\arraybackslash}m{1.8cm}>{\centering\arraybackslash}m{1.8cm}}

\toprule
    \textbf{Training data} & \textbf{Insurance (21 classes)} & \textbf{Banking \  (9 classes)} & \textbf{Finance (23 classes)}\\
    \cmidrule(lr){1-4}
    human-to-human & $92.3\pm0.5$ & $94.4\pm2.1$ & $89.7\pm0.6$\\
    transferred human-to-bot & $\bm{92.9\pm0.5}$ & $\bm{97.7\pm1.3}$ & $\bm{89.9\pm0.5}$\\
    \bottomrule
\end{tabular}}
\caption{Intent classification results in terms of F1 score. Transferring the training data (human-to-human style) to test data style (human-to-bot style) improves the test F1 score in three domains: Insurance, Banking, and Finance. The significance of difference, $p$-values for Insurance and Banking are $p<0.05$ and $p<0.01$, respectively. For Finance the improvement is non-significant.}
\label{tab:ic-results-f1-score}
\end{table}


\section{Related Works}\label{sec:related-works}
Style transfer in NLP has been studied in many variations. One line of research studied this problem as transferring to/from the style of popular novelists to/from modern English. Such as \citet{boyd2020large} used paraphrasing model for this purpose. Another variation is transferring style to a fictional movie/novel character's style as studied by \citet{han-etal-2022-meet}. Other works studied style transfer by defining style attributes and transferring text style from one attribute to another (e.g., positive/negative, informal/formal) \citep{pavlick2016empirical,rao2018dear,niu2018multi,wang2019harnessing,briakou2021ola,zhang2018style,madaan2020politeness,reif2022recipe}.  

Existing style transfer approaches make different assumptions about data availability. Certain approaches assume the availability of a lot of training data in the target style and use either a sequence-to-sequence model \citep{rao2018dear,niu2018multi,riley2021textsettr} or a controlled text generation model guided by a schema \citep{tsai2021style} or rules \citep{wang2019harnessing}. Other approaches assume the availability of zero or a small number of training examples and leverage either auto-encoders for controlled text generation such as sentiment polarity transfer and tense alteration \citep{shen2017style,mai2020plug,shen2020educating,montero2021sentence,shen2020educating} or in-context learning based on LLMs for specific attributes \citep{reif2022recipe,suzgun2022prompt,han-etal-2022-meet}.

Another line of research studied style transfer by mapping texts with different style attributes in a common latent space that is independent of the style attributes, however, preserves the semantic meaning. This approach is conceptually similar to our idea of using style-free utterances as pivots. For example, \citet{shen2017style} assumed a shared latent content distribution across different text corpora, and proposed a method that aligns the latent representations to perform style transfer. They used an adversarial discriminator to align the latent spaces of different styles. Later \citet{yang2018unsupervised} extended this idea by using language models as discriminators by addressing the instability of the error signals provided by the GAN-based discriminators. Several works have been done along the line \citep{prabhumoye2018style,gao2019structuring,madaan2020politeness} that utilized the concept of latent-space representation for style transfer. However, in a counter-study to such approaches that depend on the latent space representation for style transfer, \citet{subramanian2018multiple} showed that the assumptions related to the latent space are not necessary and are not always met in practice.

Existing works mostly ignore the context beyond a single sentence while transferring the style and rely on style attribute definitions. 
Recently, a few attempts have been made in the domain of contextual style transfer. For example, \citet{cheng2020contextual} studied style transfer of text in context where the context is defined as the paragraph where the input text appears. \citet{han-etal-2022-meet} studied style transfer in a contextualized setting where the LLMs are prompted to answer a question in the style of fictional characters. The question is used as context. However, the styles of the fictional characters are too evident and characterized by special words and other fictional characters involved in the novels or movies. In contrast, in this paper, we study style transfer in Task-Oriented Dialogues where (1) the context is the previous turns among the speakers, (2) there are only a few examples of the target style available, and (3) the style attributes are unknown and the conversation style may be a combination of many style attributes. 

Recent surveys have emphasized applications of text style transfer in domain adaptation \cite{jin2022deep}. In this paper, we take the first step towards applying style transfer to adapt training data for the downstream task of intent classification.
\section{Conclusion}\label{sec:summary}
In this paper, we study a novel problem of conversation style transfer using few-shot non-parallel examples. To solve this problem we propose a novel in-context learning approach that transfers the style of a source conversation to a target style using style-free conversations as pivots. Only a few non-parallel examples in source and target styles are needed for the purpose. We perform human and automatic evaluations to evaluate the style transfer quality for task-oriented dialogues on style strength, appropriateness, and semantic correctness. Quantitative and qualitative evaluations show that conversation style transfer yields more appropriate and semantically correct responses compared to utterance-level style transfer, which is crucial when applying to chatbot personalization. Finally, the usage of conversation style transfer for domain adaptation of training data for downstream intent classification task showed improvement in F1 score. 


\section*{Limitations}

    
    We construct styled-to-style-free parallel conversations manually using human supervision. This may be expensive to do when there are a large number of style domains. 
    An automatic measure would be ideal for this purpose and this can be an interesting future work.
    
    
    We ran our experiments only on one language, English. Various steps of the approach may be difficult to perform if style transfer is done in other languages as styles in different languages depend highly on social cultures and norms. That is mostly because most of the Large Language Models are pre-trained only on English text and may not perform well in other languages. Replicating this study in other languages may be an interesting future work.

    New LLMs of different parameter sizes have been proposed in recent times. Replicating our study with other available LLMs of different parameter sizes can be an interesting future work.

    
\section*{Ethics Statement}
In this paper, we did not annotate any new dataset rather we ran our models on publicly available datasets. The DSTC11 dataset is licensed under the Apache-2.0 License and the TWCS dataset is licensed under CC BY-NC-SA 4.0, both allow non-commercial use and distribution. The dataset references are cited and we provide detailed statistics of the dataset used. 

The examples shown in Table \ref{tab:off-the-shelf-example} are from real customer care agents from different companies and are taken from the TWCS dataset. The examples from these companies were selected only for studying the problem using real data, the authors in this paper have no connection to these companies. Note that, the identity of the individual agents is hidden in the original dataset. Hence, it does not contain any personal identification information. The signatures of names at the end of the response by the Chipotle agents from the TWCS dataset are already altered to hide the actual identity of the agents. 

We performed a human evaluation of our proposed models in this paper. We made sure that the human evaluation UIs do not impose any cognitive bias towards a specific model. We ensured that by hiding model names, shuffling orders of model outputs, and so on. We provide inter-annotator agreement scores and the detailed human evaluation process in the paper and in the appendix. Corresponding appendices are appropriately referred to in the paper.


The model descriptions and all hyper-parameter details are provided in the paper. Hence, we believe our results are reproducible. 


Any generated texts that are reported as examples in this paper are the outputs of machine learning models and do not represent the authors' or the organization's viewpoints in any way. 

Language models are pre-trained on large amounts of human-generated text. Hence, recent studies \citep{blodgett2020language,brown2020language} have discussed that there may be inherent social and human biases in these models. However, probing the increasing number of Large Language Models for biases is a separate and broad research area and falls outside the scope of our study in this paper.  

\section*{Acknowledgements}
We gratefully acknowledge Justin Sun for his help in setting up the LLMs and the members of the AWS AI Labs for providing valuable feedback on the project. We express our gratitude to the AWS AI Data Team for supporting us with the human evaluation. We would also like the thank the anonymous reviewers for their insightful comments.


\bibliography{anthology,custom}

\begin{thebibliography}{37}
\expandafter\ifx\csname natexlab\endcsname\relax\def\natexlab#1{#1}\fi

\bibitem[{Axelbrooke(2017)}]{stuart_axelbrooke_2017}
Stuart Axelbrooke. 2017.
\newblock \href {https://doi.org/10.34740/KAGGLE/DSV/8841} {Customer support on
  twitter}.

\bibitem[{Black et~al.(2022)Black, Biderman, Hallahan, Anthony, Gao, Golding,
  He, Leahy, McDonell, Phang et~al.}]{black2022gpt}
Sid Black, Stella Biderman, Eric Hallahan, Quentin Anthony, Leo Gao, Laurence
  Golding, Horace He, Connor Leahy, Kyle McDonell, Jason Phang, et~al. 2022.
\newblock Gpt-neox-20b: An open-source autoregressive language model.
\newblock \emph{Challenges \& Perspectives in Creating Large Language Models},
  page~95.

\bibitem[{Blodgett et~al.(2020)Blodgett, Barocas, Daum{\'e}~III, and
  Wallach}]{blodgett2020language}
Su~Lin Blodgett, Solon Barocas, Hal Daum{\'e}~III, and Hanna Wallach. 2020.
\newblock Language (technology) is power: A critical survey of “bias” in
  nlp.
\newblock In \emph{Proceedings of the 58th Annual Meeting of the Association
  for Computational Linguistics}, pages 5454--5476.

\bibitem[{Boyd et~al.(2020)Boyd, Puri, Shoeybi, Patwary, and
  Catanzaro}]{boyd2020large}
Alex Boyd, Raul Puri, Mohammad Shoeybi, Mostofa Patwary, and Bryan Catanzaro.
  2020.
\newblock Large scale multi-actor generative dialog modeling.
\newblock In \emph{Proceedings of the 58th Annual Meeting of the Association
  for Computational Linguistics}, pages 66--84.

\bibitem[{Briakou et~al.(2021)Briakou, Lu, Zhang, and
  Tetreault}]{briakou2021ola}
Eleftheria Briakou, Di~Lu, Ke~Zhang, and Joel Tetreault. 2021.
\newblock Ol{\'a}, bonjour, salve! xformal: A benchmark for multilingual
  formality style transfer.
\newblock In \emph{Proceedings of the 2021 Conference of the North American
  Chapter of the Association for Computational Linguistics: Human Language
  Technologies}, pages 3199--3216.

\bibitem[{Brown et~al.(2020)Brown, Mann, Ryder, Subbiah, Kaplan, Dhariwal,
  Neelakantan, Shyam, Sastry, Askell et~al.}]{brown2020language}
Tom Brown, Benjamin Mann, Nick Ryder, Melanie Subbiah, Jared~D Kaplan, Prafulla
  Dhariwal, Arvind Neelakantan, Pranav Shyam, Girish Sastry, Amanda Askell,
  et~al. 2020.
\newblock Language models are few-shot learners.
\newblock \emph{Advances in neural information processing systems},
  33:1877--1901.

\bibitem[{Cheng et~al.(2020)Cheng, Gan, Zhang, Elachqar, Li, and
  Liu}]{cheng2020contextual}
Yu~Cheng, Zhe Gan, Yizhe Zhang, Oussama Elachqar, Dianqi Li, and Jingjing Liu.
  2020.
\newblock Contextual text style transfer.
\newblock In \emph{Findings of the Association for Computational Linguistics:
  EMNLP 2020}, pages 2915--2924.

\bibitem[{Church and Hanks(1990)}]{church-hanks-1990-word}
Kenneth~Ward Church and Patrick Hanks. 1990.
\newblock \href {https://aclanthology.org/J90-1003} {Word association norms,
  mutual information, and lexicography}.
\newblock \emph{Computational Linguistics}, 16(1):22--29.

\bibitem[{Danescu-Niculescu-Mizil et~al.(2013)Danescu-Niculescu-Mizil, Sudhof,
  Jurafsky, Leskovec, and Potts}]{danescu2013computational}
Cristian Danescu-Niculescu-Mizil, Moritz Sudhof, Dan Jurafsky, Jure Leskovec,
  and Christopher Potts. 2013.
\newblock A computational approach to politeness with application to social
  factors.
\newblock In \emph{Proceedings of the 51st Annual Meeting of the Association
  for Computational Linguistics (Volume 1: Long Papers)}, pages 250--259.

\bibitem[{Gao et~al.(2018)Gao, Galley, and Li}]{gao2018neural}
Jianfeng Gao, Michel Galley, and Lihong Li. 2018.
\newblock Neural approaches to conversational ai.
\newblock In \emph{The 41st International ACM SIGIR Conference on Research \&
  Development in Information Retrieval}, pages 1371--1374.

\bibitem[{Gao et~al.(2019)Gao, Zhang, Lee, Galley, Brockett, Gao, and
  Dolan}]{gao2019structuring}
Xiang Gao, Yizhe Zhang, Sungjin Lee, Michel Galley, Chris Brockett, Jianfeng
  Gao, and Bill Dolan. 2019.
\newblock Structuring latent spaces for stylized response generation.
\newblock \emph{arXiv preprint arXiv:1909.05361}.

\bibitem[{Han et~al.(2022)Han, Kim, Yoo, Seo, Kim, Erdenee, and
  Chang}]{han-etal-2022-meet}
Seungju Han, Beomsu Kim, Jin~Yong Yoo, Seokjun Seo, Sangbum Kim, Enkhbayar
  Erdenee, and Buru Chang. 2022.
\newblock \href {https://doi.org/10.18653/v1/2022.naacl-main.377} {Meet your
  favorite character: Open-domain chatbot mimicking fictional characters with
  only a few utterances}.
\newblock In \emph{Proceedings of the 2022 Conference of the North American
  Chapter of the Association for Computational Linguistics: Human Language
  Technologies}, pages 5114--5132, Seattle, United States. Association for
  Computational Linguistics.

\bibitem[{Holtzman et~al.(2019)Holtzman, Buys, Du, Forbes, and
  Choi}]{holtzman2019curious}
Ari Holtzman, Jan Buys, Li~Du, Maxwell Forbes, and Yejin Choi. 2019.
\newblock The curious case of neural text degeneration.
\newblock In \emph{International Conference on Learning Representations}.

\bibitem[{Jin et~al.(2022)Jin, Jin, Hu, Vechtomova, and Mihalcea}]{jin2022deep}
Di~Jin, Zhijing Jin, Zhiting Hu, Olga Vechtomova, and Rada Mihalcea. 2022.
\newblock Deep learning for text style transfer: A survey.
\newblock \emph{Computational Linguistics}, 48(1):155--205.

\bibitem[{Krippendorff(2004)}]{krippendorff2004measuring}
Klaus Krippendorff. 2004.
\newblock Measuring the reliability of qualitative text analysis data.
\newblock \emph{Quality and quantity}, 38:787--800.

\bibitem[{Liu et~al.(2019)Liu, Ott, Goyal, Du, Joshi, Chen, Levy, Lewis,
  Zettlemoyer, and Stoyanov}]{liu2019roberta}
Yinhan Liu, Myle Ott, Naman Goyal, Jingfei Du, Mandar Joshi, Danqi Chen, Omer
  Levy, Mike Lewis, Luke Zettlemoyer, and Veselin Stoyanov. 2019.
\newblock Roberta: A robustly optimized bert pretraining approach.
\newblock \emph{arXiv preprint arXiv:1907.11692}.

\bibitem[{Madaan et~al.(2020)Madaan, Setlur, Parekh, Pocz{\'o}s, Neubig, Yang,
  Salakhutdinov, Black, and Prabhumoye}]{madaan2020politeness}
Aman Madaan, Amrith Setlur, Tanmay Parekh, Barnab{\'a}s Pocz{\'o}s, Graham
  Neubig, Yiming Yang, Ruslan Salakhutdinov, Alan~W Black, and Shrimai
  Prabhumoye. 2020.
\newblock Politeness transfer: A tag and generate approach.
\newblock In \emph{Proceedings of the 58th Annual Meeting of the Association
  for Computational Linguistics}, pages 1869--1881.

\bibitem[{Mai et~al.(2020)Mai, Pappas, Montero, Smith, and
  Henderson}]{mai2020plug}
Florian Mai, Nikolaos Pappas, Ivan Montero, Noah~A Smith, and James Henderson.
  2020.
\newblock Plug and play autoencoders for conditional text generation.
\newblock In \emph{Proceedings of the 2020 Conference on Empirical Methods in
  Natural Language Processing (EMNLP)}, pages 6076--6092.

\bibitem[{Montero et~al.(2021)Montero, Pappas, and Smith}]{montero2021sentence}
Ivan Montero, Nikolaos Pappas, and Noah~A Smith. 2021.
\newblock Sentence bottleneck autoencoders from transformer language models.
\newblock In \emph{Proceedings of the 2021 Conference on Empirical Methods in
  Natural Language Processing}, pages 1822--1831.

\bibitem[{Ni et~al.(2022)Ni, Young, Pandelea, Xue, and Cambria}]{ni2022recent}
Jinjie Ni, Tom Young, Vlad Pandelea, Fuzhao Xue, and Erik Cambria. 2022.
\newblock Recent advances in deep learning based dialogue systems: A systematic
  survey.
\newblock \emph{Artificial Intelligence Review}, pages 1--101.

\bibitem[{Niu et~al.(2018)Niu, Rao, and Carpuat}]{niu2018multi}
Xing Niu, Sudha Rao, and Marine Carpuat. 2018.
\newblock Multi-task neural models for translating between styles within and
  across languages.
\newblock In \emph{Proceedings of the 27th International Conference on
  Computational Linguistics}, pages 1008--1021.

\bibitem[{Pavlick and Tetreault(2016)}]{pavlick2016empirical}
Ellie Pavlick and Joel Tetreault. 2016.
\newblock An empirical analysis of formality in online communication.
\newblock \emph{Transactions of the Association for Computational Linguistics},
  4:61--74.

\bibitem[{Prabhumoye et~al.(2018)Prabhumoye, Tsvetkov, Salakhutdinov, and
  Black}]{prabhumoye2018style}
Shrimai Prabhumoye, Yulia Tsvetkov, Ruslan Salakhutdinov, and Alan~W Black.
  2018.
\newblock Style transfer through back-translation.
\newblock In \emph{Proceedings of the 56th Annual Meeting of the Association
  for Computational Linguistics (Volume 1: Long Papers)}, pages 866--876.

\bibitem[{Rao and Tetreault(2018)}]{rao2018dear}
Sudha Rao and Joel Tetreault. 2018.
\newblock Dear sir or madam, may i introduce the gyafc dataset: Corpus,
  benchmarks and metrics for formality style transfer.
\newblock In \emph{Proceedings of the 2018 Conference of the North American
  Chapter of the Association for Computational Linguistics: Human Language
  Technologies, Volume 1 (Long Papers)}, pages 129--140.

\bibitem[{Reif et~al.(2022)Reif, Ippolito, Yuan, Coenen, Callison-Burch, and
  Wei}]{reif2022recipe}
Emily Reif, Daphne Ippolito, Ann Yuan, Andy Coenen, Chris Callison-Burch, and
  Jason Wei. 2022.
\newblock A recipe for arbitrary text style transfer with large language
  models.
\newblock In \emph{Proceedings of the 60th Annual Meeting of the Association
  for Computational Linguistics (Volume 2: Short Papers)}, pages 837--848.

\bibitem[{Reimers and Gurevych(2019)}]{reimers2019sentence}
Nils Reimers and Iryna Gurevych. 2019.
\newblock Sentence-bert: Sentence embeddings using siamese bert-networks.
\newblock In \emph{Proceedings of the 2019 Conference on Empirical Methods in
  Natural Language Processing and the 9th International Joint Conference on
  Natural Language Processing (EMNLP-IJCNLP)}, pages 3982--3992.

\bibitem[{Riley et~al.(2021)Riley, Constant, Guo, Kumar, Uthus, and
  Parekh}]{riley2021textsettr}
Parker Riley, Noah Constant, Mandy Guo, Girish Kumar, David~C Uthus, and Zarana
  Parekh. 2021.
\newblock Textsettr: Few-shot text style extraction and tunable targeted
  restyling.
\newblock In \emph{Proceedings of the 59th Annual Meeting of the Association
  for Computational Linguistics and the 11th International Joint Conference on
  Natural Language Processing (Volume 1: Long Papers)}, pages 3786--3800.

\bibitem[{Shen et~al.(2017)Shen, Lei, Barzilay, and Jaakkola}]{shen2017style}
Tianxiao Shen, Tao Lei, Regina Barzilay, and Tommi Jaakkola. 2017.
\newblock Style transfer from non-parallel text by cross-alignment.
\newblock \emph{Advances in neural information processing systems}, 30.

\bibitem[{Shen et~al.(2020)Shen, Mueller, Barzilay, and
  Jaakkola}]{shen2020educating}
Tianxiao Shen, Jonas Mueller, Regina Barzilay, and Tommi Jaakkola. 2020.
\newblock Educating text autoencoders: Latent representation guidance via
  denoising.
\newblock In \emph{International conference on machine learning}, pages
  8719--8729. PMLR.

\bibitem[{Subramanian et~al.(2018)Subramanian, Lample, Smith, Denoyer, Ranzato,
  and Boureau}]{subramanian2018multiple}
Sandeep Subramanian, Guillaume Lample, Eric~Michael Smith, Ludovic Denoyer,
  Marc'Aurelio Ranzato, and Y-Lan Boureau. 2018.
\newblock Multiple-attribute text style transfer.
\newblock \emph{arXiv preprint arXiv:1811.00552}.

\bibitem[{Suzgun et~al.(2022)Suzgun, Melas-Kyriazi, and
  Jurafsky}]{suzgun2022prompt}
Mirac Suzgun, Luke Melas-Kyriazi, and Dan Jurafsky. 2022.
\newblock Prompt-and-rerank: A method for zero-shot and few-shot arbitrary
  textual style transfer with small language models.
\newblock In \emph{Proceedings of the 2022 Conference on Empirical Methods in
  Natural Language Processing}, pages 2195--2222.

\bibitem[{Tsai et~al.(2021)Tsai, Oraby, Perera, Kao, Du, Narayan-Chen, Chung,
  and Hakkani-Tur}]{tsai2021style}
Alicia Tsai, Shereen Oraby, Vittorio Perera, Jiun-Yu Kao, Yuheng Du, Anjali
  Narayan-Chen, Tagyoung Chung, and Dilek Hakkani-Tur. 2021.
\newblock Style control for schema-guided natural language generation.
\newblock In \emph{Proceedings of the 3rd Workshop on Natural Language
  Processing for Conversational AI}, pages 228--242.

\bibitem[{Wang et~al.(2019)Wang, Wu, Mou, Li, and Chao}]{wang2019harnessing}
Yunli Wang, Yu~Wu, Lili Mou, Zhoujun Li, and Wenhan Chao. 2019.
\newblock Harnessing pre-trained neural networks with rules for formality style
  transfer.
\newblock In \emph{Proceedings of the 2019 Conference on Empirical Methods in
  Natural Language Processing and the 9th International Joint Conference on
  Natural Language Processing (EMNLP-IJCNLP)}, pages 3573--3578.

\bibitem[{Yang et~al.(2018)Yang, Hu, Dyer, Xing, and
  Berg-Kirkpatrick}]{yang2018unsupervised}
Zichao Yang, Zhiting Hu, Chris Dyer, Eric~P Xing, and Taylor Berg-Kirkpatrick.
  2018.
\newblock Unsupervised text style transfer using language models as
  discriminators.
\newblock \emph{Advances in Neural Information Processing Systems}, 31.

\bibitem[{Zar(2005)}]{zar2005spearman}
Jerrold~H Zar. 2005.
\newblock Spearman rank correlation.
\newblock \emph{Encyclopedia of biostatistics}, 7.

\bibitem[{Zhang et~al.(2020)Zhang, Takanobu, Zhu, Huang, and
  Zhu}]{zhang2020recent}
Zheng Zhang, Ryuichi Takanobu, Qi~Zhu, MinLie Huang, and XiaoYan Zhu. 2020.
\newblock Recent advances and challenges in task-oriented dialog systems.
\newblock \emph{Science China Technological Sciences}, 63(10):2011--2027.

\bibitem[{Zhang et~al.(2018)Zhang, Ren, Liu, Wang, Chen, Li, Zhou, and
  Chen}]{zhang2018style}
Zhirui Zhang, Shuo Ren, Shujie Liu, Jianyong Wang, Peng Chen, Mu~Li, Ming Zhou,
  and Enhong Chen. 2018.
\newblock Style transfer as unsupervised machine translation.
\newblock \emph{arXiv preprint arXiv:1808.07894}.

\end{thebibliography}
\bibliographystyle{acl_natbib}

\newpage
\appendix

\section{Prompting}\label{appendix:prompting}
\subsection{Prompt Structure}
\begin{figure*}[h]
  \centering
  \includegraphics[width=1\textwidth]{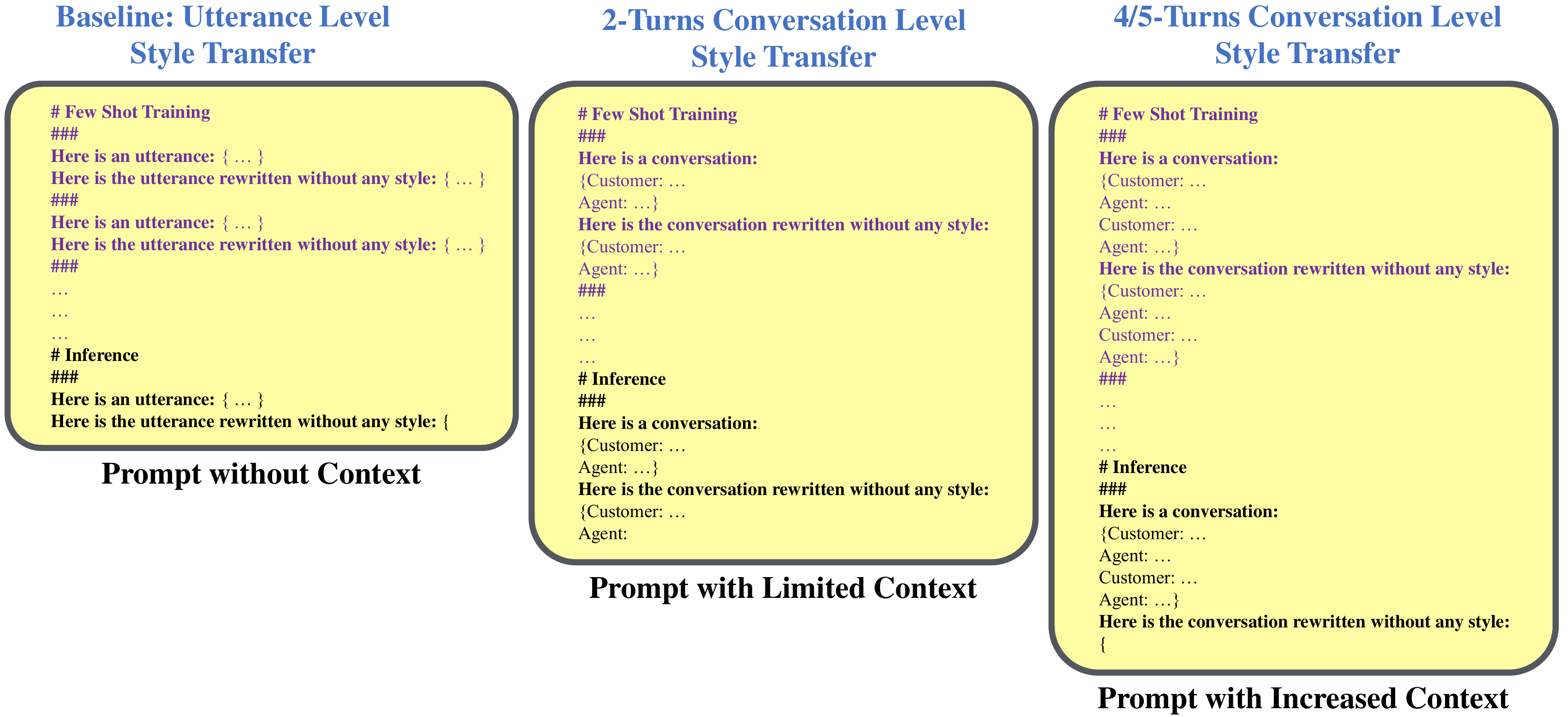}
  \caption{Prompt structure for transferring a source conversation to a style free conversation.}
  \label{fig:appendix-prompt-structure-to-style-free}
\end{figure*}

\begin{figure*}[h]
  \centering
  \includegraphics[width=1\textwidth]{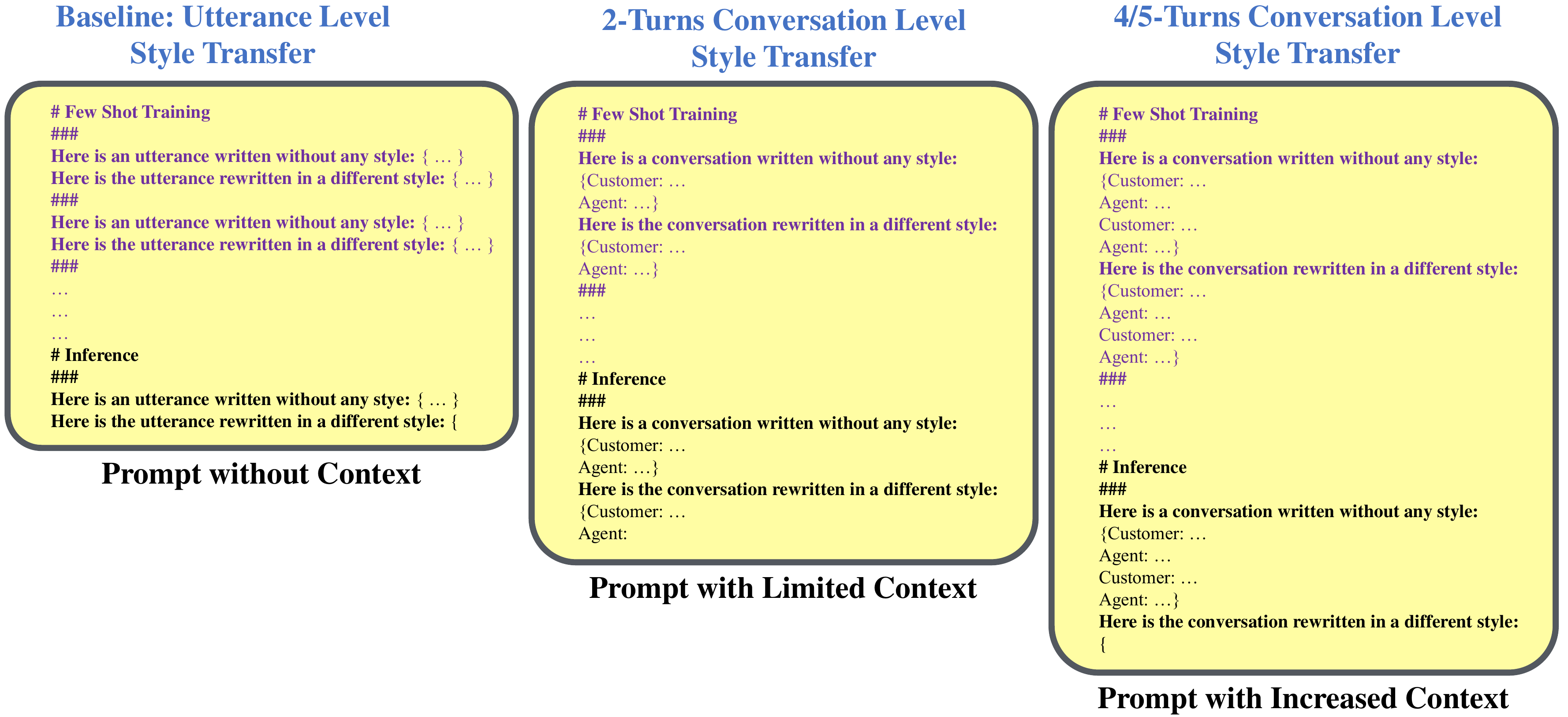}
  \caption{Prompt structure for transferring a style free conversation to the target style.}
  \label{fig:appendix-prompt-structure-to-target-style}
\end{figure*}

The structure of prompts for various versions of our model for converting a source conversation to style free conversation are shown in Figure \ref{fig:appendix-prompt-structure-to-style-free}. The prompt structures for converting style free conversation to the target style are shown in Figure \ref{fig:appendix-prompt-structure-to-target-style}.

\subsection{Prompt Example}
\begin{figure*}[h]
  \centering
  \includegraphics[width=1\textwidth]{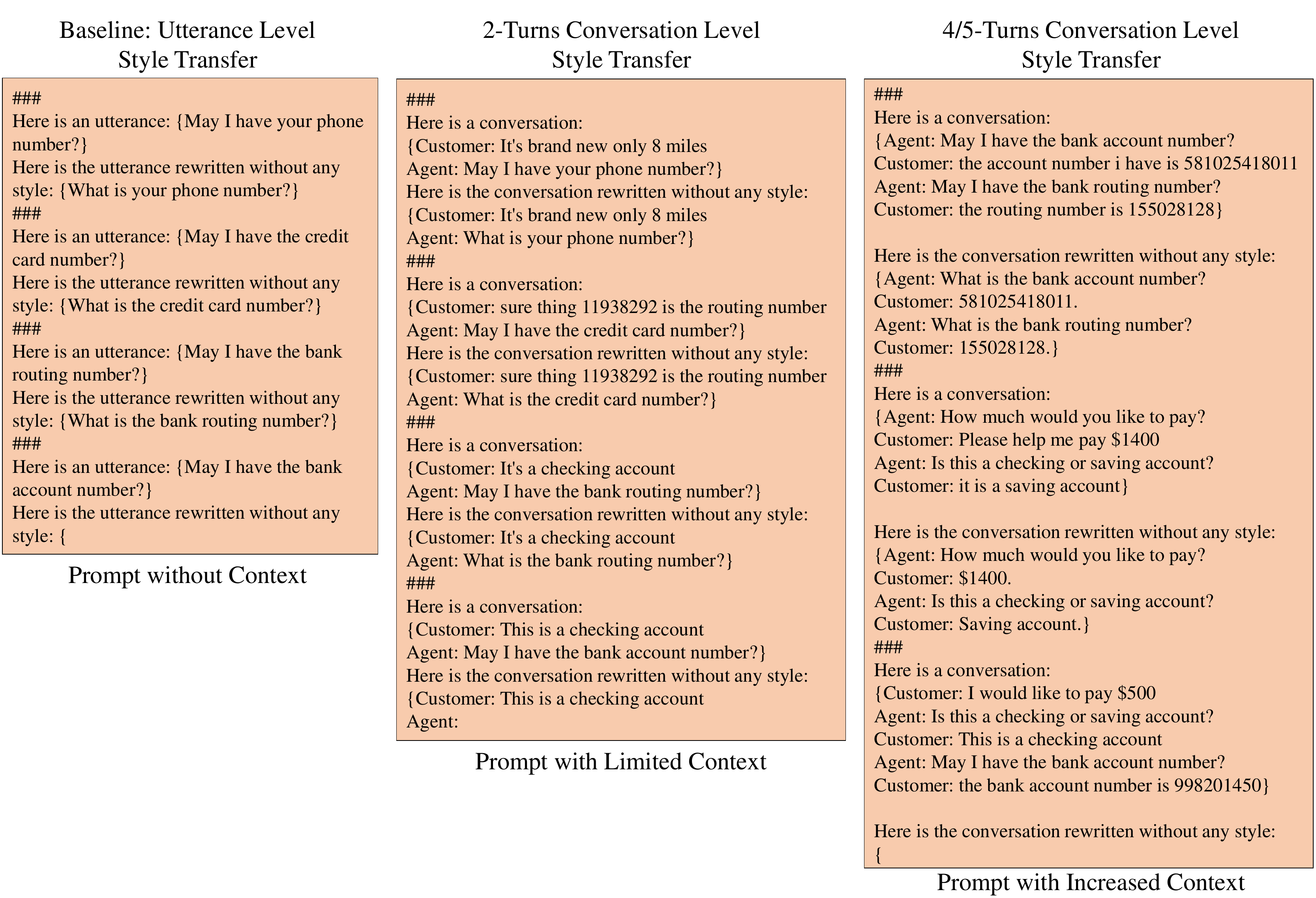}
  \caption{Prompt examples for transferring a source conversation (in chatbot agent style, $B$) to a style free conversation using various versions of our model. For simplicity, 3-shot, 3-shot, and 2-shot prompts are shown in case of utterance level style transfer, 2-turns conversation level style transfer and 4/5-turns conversation level style transfer, respectively.}
  \label{fig:appendix-prompt-example-to-style-free}
\end{figure*}

\begin{figure*}[h]
  \centering
  \includegraphics[width=1\textwidth]{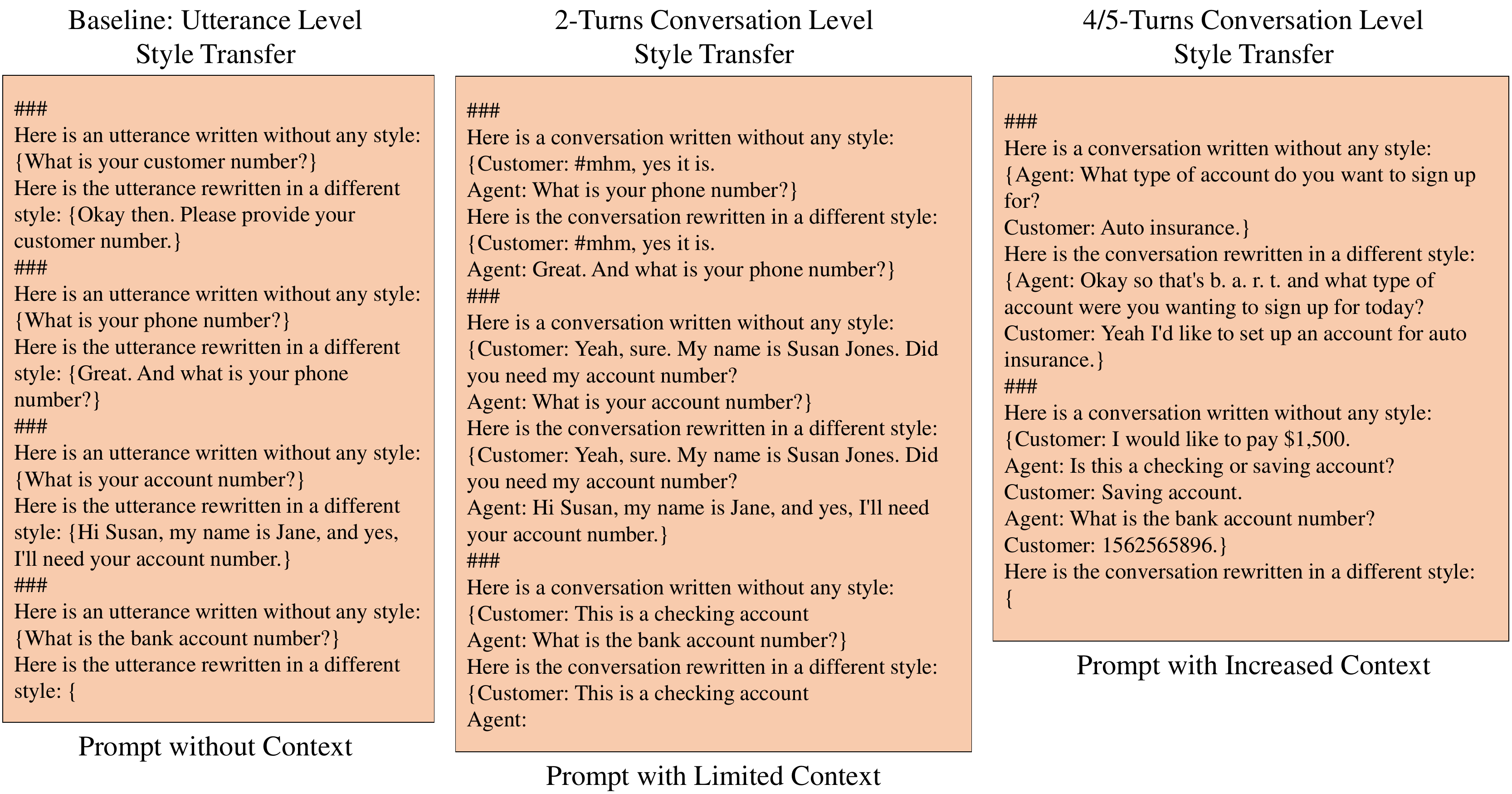}
  \caption{Prompt structure for transferring a style free conversation to the target style (in human agent style, $H_1$) using various versions of our model. For simplicity, 3, 2, 1 shots prompts are shown in case of utterance level style transfer, 2-turns conversation level style transfer and 4/5-turns conversation level style transfer, respectively.}
  \label{fig:appendix-prompt-example-to-target-style}
\end{figure*}

Examples for all types of prompt structures (as shown in Figure \ref{fig:appendix-prompt-structure-to-style-free} and Figure \ref{fig:appendix-prompt-structure-to-target-style}) are shown in Figures \ref{fig:appendix-prompt-example-to-style-free} and \ref{fig:appendix-prompt-example-to-target-style}.  

\section{Example Conversations from Various Domains}\label{appendix:example-conversations}
\begin{figure*}[h]
    \begin{center}
        \includegraphics[width=1\textwidth]{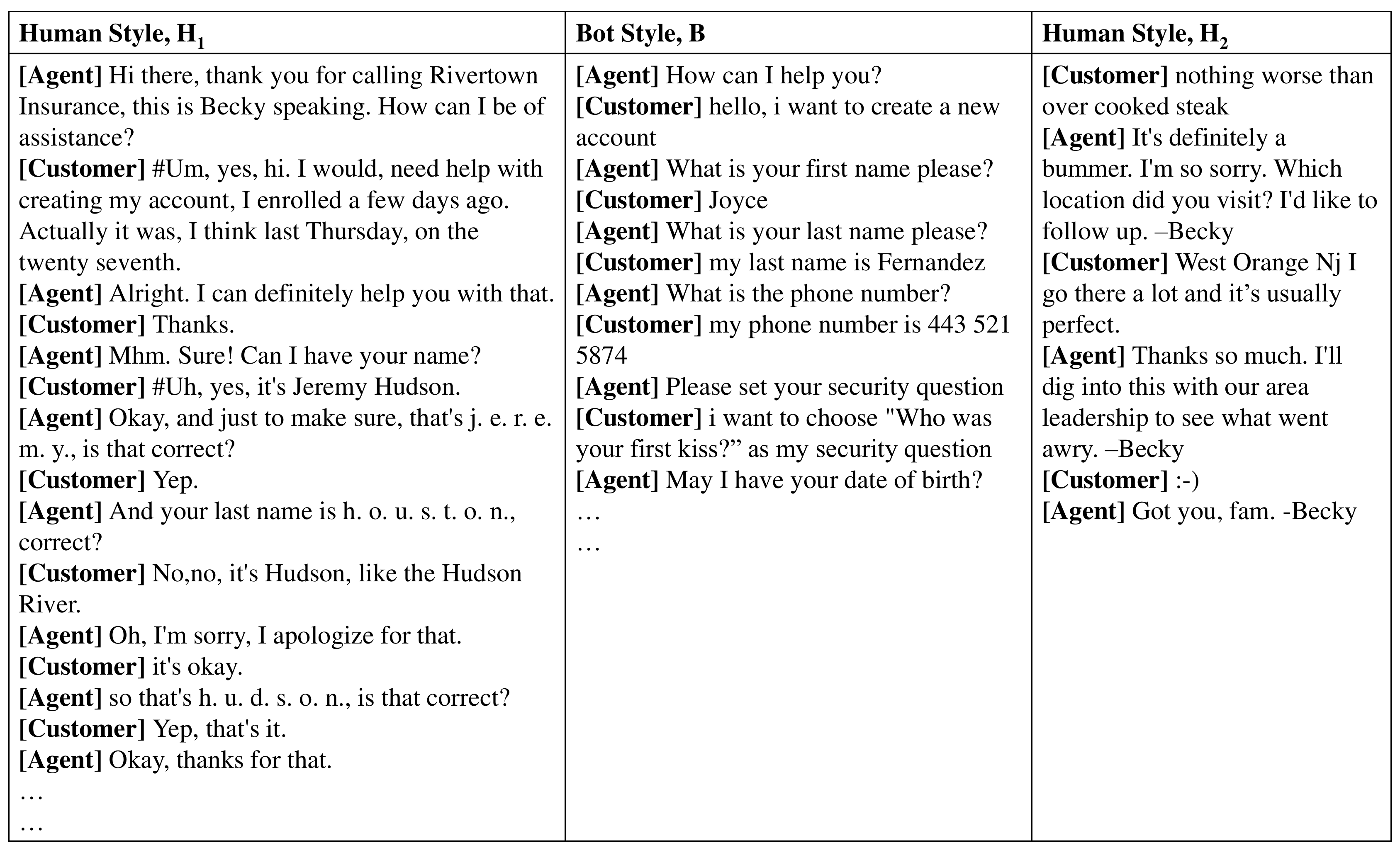}
    \caption{Example conversations in styles $H_1, B, H_2$. These three styles are holistically different. Some properties of the human styles are that they are conversational, sometimes formal, and sometimes casual and friendly. For example, the human style, $H_2$ is informal and friendly while the other human style, $H_1$ is formal while both of these two human styles are conversational. In human style $H_2$, agents sign their names at the end of a response preceded by a hyphen. In the other human style, $H_1$, this stylistic property is not observed. Some observed properties of the bot style are crispness and to-the-point while not being informal. The conversations in human style $H_1$ and bot style $B$ presented in this table, are on the same situation showing the holistic difference between these two styles. Note that, the other human style $H_2$ is from a different domain, hence, a conversation on a similar situation could not be found.}
    \label{fig:qualitative-style-examples}
    \end{center}
\end{figure*}

Example conversations for chatbot style (referred to as $B$) and the two human styles $H_1, H_2$ are shown in Figure \ref{fig:qualitative-style-examples}.

\subsection{PMI-based Style Indicator Lemma Identification}\label{appendix:pmi-lemma-procedure}
For the identification of style indicator lemmas for each style domain, we use a Pointwise Mutual Information (PMI) \cite{church-hanks-1990-word} based approach. We first take all of the agent utterances from each style domain and lemmatize each word used by the agents using the spaCy Python library. We ignore all punctuations and stopwords. Then for a lemma, $w$ we calculate the pointwise mutual information (PMI) with a style domain $t$, $I(w, t)$ using the following formula.
\begin{align*}
     I(w,t)=\operatorname{log}\frac{P(w|t)}{P(w)}
\end{align*}{}
Where $P(w|t)$ is computed by taking all lemmas used in style $t$ and computing $\frac{count(w)}{count(all-lemmas)}$ and similarly, $P(w)$ is computed by counting lemma $w$ over the set of utterances in all styles. Now, we rank lemmas for each style domain based on their PMI scores. To remove topic-specific lemmas and rarely used lemmas, we ignore lemmas that are used in more than $10\%$ of the agent utterances in each style domain and used less than $0.5\%, 0.3\%, 0.3\%$ of the time in case of styles $H_1, B, H_2$, respectively. The top 300 high PMI lemmas for each style domain are reported in Table \ref{tab:style-lexicon-all}. Hand-picked style indicator lemmas from this top 300 list are shown in Table \ref{tab:style-lexicon}.

\begin{table*}[t!]
\centering
\resizebox{2\columnwidth}{!}{%
\begin{tabular}
{>{\arraybackslash}m{0.9cm}>{\arraybackslash}m{20cm}}

    \toprule
    \textbf{Styles} & \textbf{High PMI style indicator lemmas (written in descending order of the PMI scores)}\\
    \cmidrule(lr){1-2}
    $H_1$   &  verify, receive, mister, payment, course, moment, correct, alright, sorry, nineteen, ready, process, agent, assist, due, anything, else, high, social, file, auto, claim, actually, website, thirty, dot, com, got, pull, mother, maiden, dollar, twenty, premium, mail, digit, ahead, rest, kindly, bye, monthly, second, choose, complete, proceed, basic, preferred, coverage, rate, quote, spell, life, petcare, eighty, offer, month, system, fifty, mhm, cancel, uh, log, um, survey, worry, huh, password, reset, morning, pleasure, easy, confirmation, sir, confirm, goodbye, fine, cost, ok, afternoon, number, yes, information, name, problem, great, may, need, today, understand, call, help, day, miss, yeah, take, also, add, update, rivertown, insurance, perfect, hold, oh, minute, say, well, enjoy, year, full, end, customer, find, thing, option, mean, go, send, good, bill, sure, care, thank, look, nice, change, pet, long, set, cover, see, provide, glad, use, get, account, still, mind, right, hello, contact, pay, way, online, think, link, back, tell, let, security, hope, definitely, next, speak, damage, come, start, happy, check, service, able, question, home, time, ask, email, policy, want, welcome, work, know, give, sound, billing, plan, make, try, first, hear, last, answer, please, detail, new, phone, car, birth, card, address, date, much, code, zip, happen, accident, type, credit\\
    \cmidrule(lr){1-2}
    $B$     & apartment, street, routing, relationship, state, live, mileage, complex, unit, rough, estimation, value, property, insure, insured, person, pass, cause, death, city, ssn, condo, frequency, period, checking, save, bank, expiration, vehicle, gender, tobacco, consumption, level, height, weight, amount, preexist, condition, driver, license, incident, enroll, dependent, health, breed, age, weigh, group, additional, purpose, doctor, total, enrol, encounter, model, cvv, issue, charge, away, credit, type, accident, zip, code, date, address, card, much, birth, car, visit, phone, new, answer, happen, last, first, billing, plan, please, give, policy, home, email, question, damage, start, detail, security, service, welcome, want, make, pay, hello, time, account, cover, set, check, pet, provide, bill, change, customer, year, know, insurance, rivertown\\
    \cmidrule(lr){1-2}
    $H_2$   & every, though, note, touch, team, cool, lot, hopefully, never, remove, barbacoa, bring, ca, bean, 3, chat, manager, reach, steak, order, enough, feel, free, restaurant, really, inconvenience, always, feedback, management, loop, hand, exact, apology, leave, request, ounce, unprocessed, queso, wow, totally, particular, rice, keep, week, playlist, fan, hit, seem, improve, cheese, put, line, cs, black, taco, asap, item, odd, quick, frustrating, salsa, guest, unfortunately, stay, tuned, hey, sign, disappointing, meat, awesome, case, troubling, standard, specific, tortilla, guy, side, stop, solid, choice, man, already, foil, shoot, late, shortly, wrong, fresh, kind, wish, lunch, least, bad, usually, hour, little, dig, hesitate, sofritas, dinner, luck,future, double, gon, na, eat, place, far, ta, depend, suggestion, special, word, ah, gotcha, early, select, veggie, open, concern, friend, write, share, yet, amend, sad, serve, love, follow, menu, ingredient, chip, guac, burritos, meal, bowl, dm, soon, location, appreciate, area, trouble, portion, chicken, chipotle, different, people, bag, leadership, talk, dj, food, message, burrito, ever, real, fix, close, recipe, maybe, list, hang, someone, experience, info, something, bummer, wait, extra, hard,app, field, leaders, visit, away, charge, issue, hear, try, sound, happen, work, know, ask, able, happy, come, detail, speak,make, next, definitely, hope, let, tell, back, link, think, much, online, way, contact, right, mind, still, get, use, glad, check, see, time, welcome, long, want, nice, look, thank, care, sure, good, send, go, please, mean, option, thing, find, end, full, service, enjoy, well, say, minute, oh, hold, perfect, give, update, add, also, take, yeah, miss, day, help, call, new, understand, address, change, today, provide, start, need, phone, may, great, card, email, problem, date, customer, name, information, plan, yes, question, first, last, number\\
    \bottomrule
\end{tabular}}

\caption{High PMI lemmas for each style domain. Bots ($B$) do not use many non-topic-specific words. Mostly formal words are used in human style $H_1$ and many informal and friendly words (e.g., bummer) are used in human style $H_2$.}

\label{tab:style-lexicon-all}
\end{table*}

\subsection{Construction of parallel style free conversations using human supervision}\label{appendix:creation-of-few-shot-examples}
\begin{figure*}[h]
  \centering
  \includegraphics[width=1\textwidth]{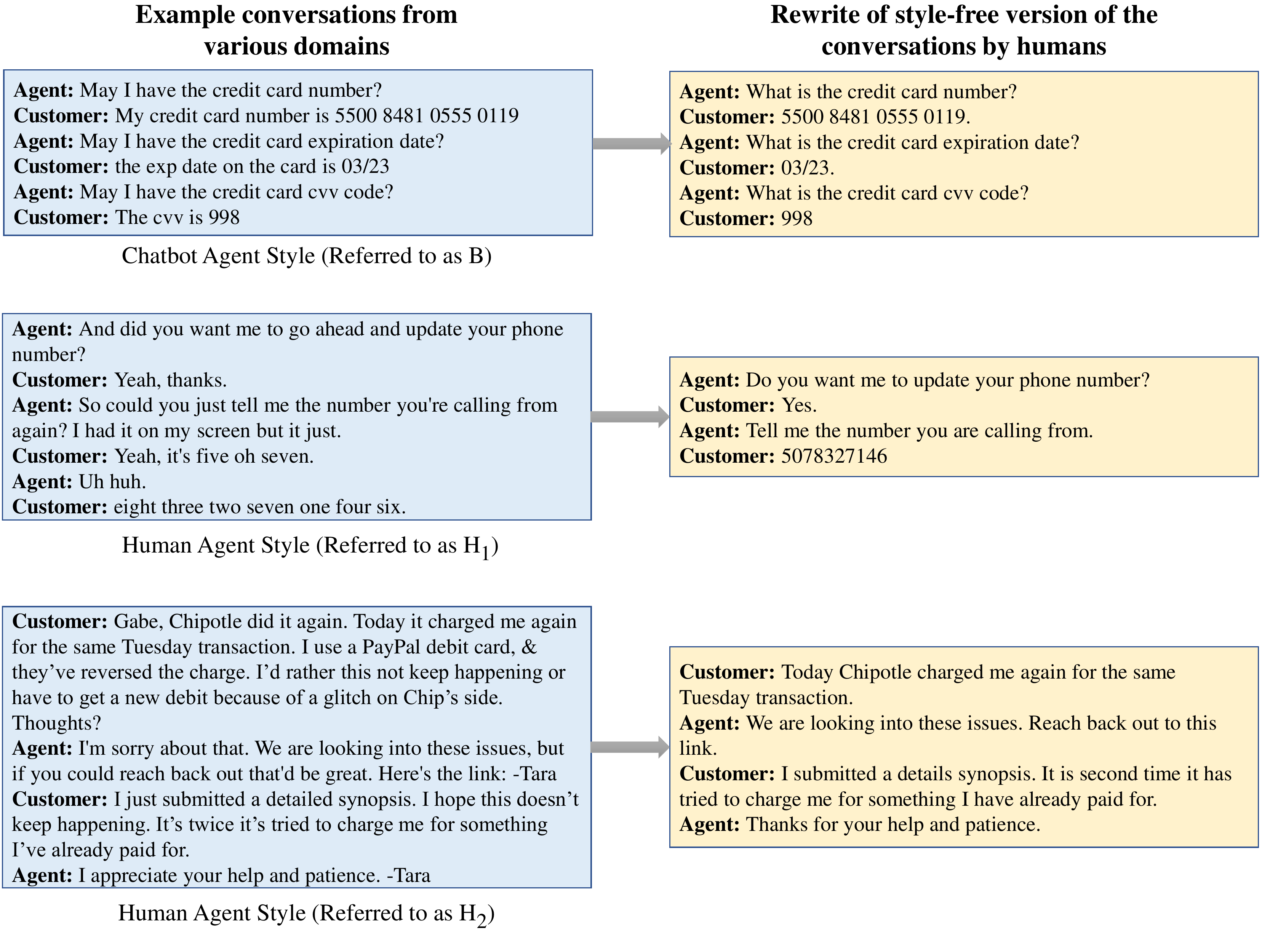}
  \caption{Example conversations from three domains ($B, H_1, H_2$) are shown in the left hand side. Human annotated style-free versions of the corresponding conversations are shown in the right hand side. This parallel data is used for in-context learning. Here, $H_1, H_2, B$ refer to human style from DSTC11 dataset, style of Chipotle agents (from TWCS dataset), and Chatbot style from DSTC11 dataset, respectively.}
  \label{fig:examples-from-various-domains}
\end{figure*}

A human annotator was presented with 5-7 conversations from each of the style domains ($B, H_1, H_2$) and they were asked to rewrite those conversations in a style-free form. One parallel style-free example per style domain written by the human annotator is shown on the right-hand side of Figure \ref{fig:examples-from-various-domains}. The human annotator is a researcher in NLP and it took approximately 5 minutes for them to rewrite a 10-12 turns conversation in a style-free format. These style-free parallel conversations are used for in-context learning as described in Section \ref{sec:modeling}. The statistics of the annotated few shot examples per style domain are shown in Table \ref{tab:few-shot-dataset-summary}.

\begin{table}
\centering
\resizebox{1\columnwidth}{!}{%
\begin{tabular}
{>{\arraybackslash}m{1cm}>{\centering\arraybackslash}m{1.3cm}>{\centering\arraybackslash}m{1.3cm}>{\centering\arraybackslash}m{1.3cm}|>{\centering\arraybackslash}m{1.3cm}>{\centering\arraybackslash}m{1.3cm}>{\centering\arraybackslash}m{1.3cm}}

\toprule
    \multicolumn{1}{c}{} & \multicolumn{3}{c}{\textbf{Utterance and 2-turns conv.}} &  \multicolumn{3}{c}{\textbf{4/5-turns conv.}} \\
    \cmidrule(lr){2-4}\cmidrule(lr){5-7}
    \textbf{\makecell[l]{Styles}} & \textbf{\# convo.} & \textbf{\# all utt.} & \textbf{\# agent utt.} & \textbf{\# segment} & \textbf{\# all utt.} & \textbf{\# agent utt.}\\
   \cmidrule(lr){1-1}\cmidrule(lr){2-4}\cmidrule(lr){5-7}
    $H_1$ & 5 & 261 & 131 & 5 & 287 & 144\\
    $H_2$ & 7 & 54 & 24 & 5 & 42 & 19\\
    $B$ & 5 & 100 & 50  & 7 & 124 & 62\\
    \bottomrule
\end{tabular}}
\caption{Manually created few-shot examples summary. The data is used for in-context learning as described in Section \ref{sec:modeling}. Here, $H_1, H_2, B$ refer to human style from DSTC11 dataset, style of Chipotle agents (from TWCS dataset), and Chatbot style from DSTC11 dataset, respectively.}

\label{tab:few-shot-dataset-summary}
\end{table}

\section{Ablation Study}\label{appendix:ablation-study}
We perform ablation study to select number of shots and compare the effect of dynamic prompt selection. We experiment with 5, 10, 20 shot training for utterance level style transfer and 2-turns conversation level style transfer. Because of the limit of tokens in prompts we experiment with 4, 8 shot training for 4/5-turns conversation level style transfer. Note that with 4/5-turns context each training example contains many more tokens. In the cases of transferring to the second human style $H_2$, 20 shot training is not supported because of the prompt limit and the conversations in this style being more conversational and greater in length. We measure the effectiveness of the number of training examples and prompt selection techniques by the automatically measured style strength of the target style after style transfer. We run this ablation study on the validation dataset shown in Table \ref{tab:dataset-summary} and use GPT-NeoX as the base LLM as it is cheaper to use compared to Bigscience-Bloom. The results are shown in Table \ref{tab:appendix-ablation}. It can be seen that dynamic prompt selection outperforms random prompt selection in all of the cases. The optimum number of shots for utterance level style transfer and 2-turns conversation level style transfer is 10 and for 4/5-turns conversation level style transfer it is 8.

\begin{table*}[ht!]
\centering

\resizebox{2.1\columnwidth}{!}{%
\begin{tabular}
{>{\arraybackslash}m{1.5cm}>{\arraybackslash}m{1.6cm}>{\centering\arraybackslash}m{2cm}|>{\centering\arraybackslash}m{2cm}>{\centering\arraybackslash}m{2cm}|>{\centering\arraybackslash}m{2cm}>{\centering\arraybackslash}m{2cm}|>{\centering\arraybackslash}m{2cm}>{\centering\arraybackslash}m{2cm}}
\toprule
    \multicolumn{1}{c}{} & \multicolumn{1}{c}{} & \multicolumn{1}{c}{} & \multicolumn{6}{c}{\textbf{Target style strength after style transfer}} \\
    \cmidrule(lr){4-9}
    &  &  & \multicolumn{2}{c}{\textbf{5 shots prompting}} & \multicolumn{2}{c}{\textbf{10 shots prompting}} & \multicolumn{2}{c}{\textbf{20 shots prompting}}\\
    \cmidrule(lr){4-5}\cmidrule(lr){6-7}\cmidrule(lr){8-9}
    \textbf{Models} & \textbf{\makecell[c]{Style\\directions}} & \textbf{Target style strength before style transfer} & \textbf{Random prompt selection} & \textbf{Dynamic prompt selection} & \textbf{Random prompt selection} & \textbf{Dynamic prompt selection} & \textbf{Random prompt selection} & \textbf{Dynamic prompt selection}\\
    \cmidrule(lr){1-9}
    \multirow{5}{*}{\textbf{\makecell[l]{Utterance\\level style\\transfer}}}
    & $H_1 \rightarrow B$    & 0.010 (0.1) & 0.114 (0.3) & 0.148 (0.3) & 0.077 (0.2) & 0.150 (0.3) & 0.085 (0.3) & 0.133 (0.3)\\
    & $H_1 \rightarrow H_2$   & 0.112 (0.3) & 0.198 (0.3) & 0.239 (0.4) & 0.182 (0.3) & 0.215 (0.3) & 0.191 (0.3) & 0.225 (0.3)\\
    & $B \rightarrow H_1$   & 0.001 (0) & 0.254 (0.4) & 0.451 (0.5) & 0.411 (0.5) & 0.556 (0.5) & 0.288 (0.4) & 0.389 (0.5)\\
    & $B \rightarrow H_2$   & 0 (0)     & 0.241 (0.4) & 0.523 (0.5) & 0.337 (0.5) & 0.671 (0.4) & 0.361 (0.5) & 0.477 (0.5)\\
    \cmidrule(lr){2-9}
    & Average           & 0.031 & 0.202 & 0.340 & 0.252 & 0.398 & 0.231 & 0.306\\
    \cmidrule(lr){1-9}
    
    \multirow{5}{*}{\textbf{\makecell[l]{2-turns\\conv. level\\style tran.}}}
    & $H_1 \rightarrow B$    & 0.010 (0.1) & 0.046 (0.2) & 0.121 (0.3) & 0.045 (0.2) & 0.119 (0.3) & 0.061 (0.2) & 0.109 (0.3)\\
    & $H_1 \rightarrow H_2$   & 0.112 (0.3) & 0.160 (0.3) & 0.173 (0.3) & 0.165 (0.3) & 0.199 (0.3) & N/S   & N/S\\
    & $B \rightarrow H_1$   & 0.001 (0) & 0.115 (0.3) & 0.410 (0.5) & 0.101 (0.3) & 0.399 (0.5) & 0.147 (0.3) & 0.476 (0.5)\\
    & $B \rightarrow H_2$   & 0 (0)     & 0.012 (0.1) & 0.052 (0.2) & 0.062 (0.2) & 0.113 (0.3) & N/S   & N/S\\
    \cmidrule(lr){2-9}
    & Average           & 0.031 & 0.083 & 0.189 & 0.093 & 0.208 & 0.104 & 0.293\\
    \cmidrule(lr){1-9}
    
    & & & \multicolumn{2}{c}{\textbf{4 shots prompting}} & \multicolumn{2}{c}{\textbf{8 shots prompting}} & &\\
    \cmidrule(lr){4-5}\cmidrule(lr){6-7}
    \multirow{5}{*}{\textbf{\makecell[l]{4/5-turns\\conv. level\\style tran.}}}
    & $H_1 \rightarrow B$    & 0.01 (0.1) & 0.072 (0.3) & 0.162 (0.4) & 0.1 (0.3) & 0.16 (0.4) & &\\
    & $H_1 \rightarrow H_2$   & 0.112 (0.3) & 0.170 (0.3) & 0.171 (0.3) & 0.165 (0.3) & 0.173 (0.3) & &\\
    & $B \rightarrow H_1$   & 0.001 (0) & 0.258 (0.4) & 0.392 (0.5) & 0.291 (0.5) & 0.42 (0.5) & &\\
    & $B \rightarrow H_2$   & 0 (0)     & 0.13 (0.3) & 0.068 (0.3) & 0.058 (0.2) & 0.11 (0.3) & &\\
    \cmidrule(lr){2-7}
    & Average           & 0.031 & 0.158 & 0.198 & 0.154 & 0.216 & &\\
    \bottomrule
\end{tabular}}
\caption{Ablation study for selecting number of shots and prompt selection method. Here, "N/S" means "Not Supported" because of token limit in prompt. GPT-NeoX was used as the base LLM in this ablation study. Dynamic prompt selection technique outperforms random prompt selection in all of the cases. The optimal number of shots for utterance level style transfer, 2-turns conversation level style transfer, and 4/5-turns conversation level style transfer are 10, 10, and 8 respectively.}\label{tab:appendix-ablation}
\end{table*}

\section{Human Evaluation}\label{appendix:human-evaluation}
\subsection{Data Selection for Human Evaluation}
Our goal with human evaluation is to compare different models. We used the test dataset described in Table \ref{tab:dataset-summary} for human evaluation. Note that the same conversation segments are used to evaluate various versions of our model and the baseline using GPT-NeoX and Bloom as LLMs. We evaluate only agent responses and we apply two types of filtering step on these datasets before human evaluation.

\textbf{Filtering Step 1:} When doing style transfer at 4/5-turn conversation level, it may result in non-parallel conversation compared to the source conversation because of turn reduction by the model. To match the non-parallel utterances with the source utterances, we rank the style transferred utterances based on their semantic similarity with the source utterances and pick the one with the highest similarity. We discard any style transferred utterance that has the highest semantic similarity of less than 0.2. Looking manually at those utterances it was observed that those were unrelated utterances generated by the LLMs. 

\textbf{Filtering Step 2:} We filtered out all agent responses where none of the models (including the baseline) changes the source agent utterances or when the style transferred versions were the same from all models.

Application of the above two filtering steps resulted in $100$+ agent utterances in each style direction. We perform the human evaluation on this filtered set. The statistics of the data after each filtering step is shown in Table \ref{tab:dataset-for-human-evaluation}.

\begin{table*}
\centering
\resizebox{2\columnwidth}{!}{%
\begin{tabular}
{>{\arraybackslash}m{2cm}>{\centering\arraybackslash}m{2cm}>{\centering\arraybackslash}m{2cm}>{\centering\arraybackslash}m{2.5cm}>{\centering\arraybackslash}m{3cm}>{\centering\arraybackslash}m{2cm}>{\centering\arraybackslash}m{2cm}>{\centering\arraybackslash}m{2.5cm}>{\centering\arraybackslash}m{3cm}}

\toprule
    \multicolumn{1}{c}{} & \multicolumn{4}{c}{\textbf{\textsc{GPT-NeoX (20B)}}} &  \multicolumn{4}{c}{\textbf{\textsc{Bigscience-Bloom (176B)}}} \\
    \cmidrule(lr){2-5}\cmidrule(lr){6-9}
    \textbf{\makecell[c]{Style\\Directions}} & \textbf{No of segments} & \textbf{No of agent utterances} & \textbf{No of agent utterances after filtering step-1}  & \textbf{No of agent utterances after filtering step-1 \& filtering step-2} & \textbf{No of segments} & \textbf{No of agent utterances} & \textbf{No of agent utterances after filtering step-1}  & \textbf{No of agent utterances after filtering step-1 \& filtering step-2} \\
   \cmidrule(lr){1-1}\cmidrule(lr){2-5}\cmidrule(lr){6-9}
    $H_1 \rightarrow B$ & 65 & 164 & 135 & 113 & 65 & 164 & 117 & 116\\
    $H_1 \rightarrow H_2$ & 65 & 166 & 141 & 140 & 65 & 166 & 115 & 113\\
    $B \rightarrow H_1$ & 65 & 152 & 139 & 102 & 65 & 152 & 128 & 123\\
    $B \rightarrow H_2$ & 65 & 152 & 134 & 134 & 65 & 152 & 129 & 125\\
    \bottomrule
\end{tabular}}
\caption{Dataset statistics for human evaluation.}\label{tab:dataset-for-human-evaluation}
\end{table*}

\subsection{Human Evaluation Settings} 
\begin{figure*}[h!]
  \centering
  \includegraphics[width=1\textwidth]{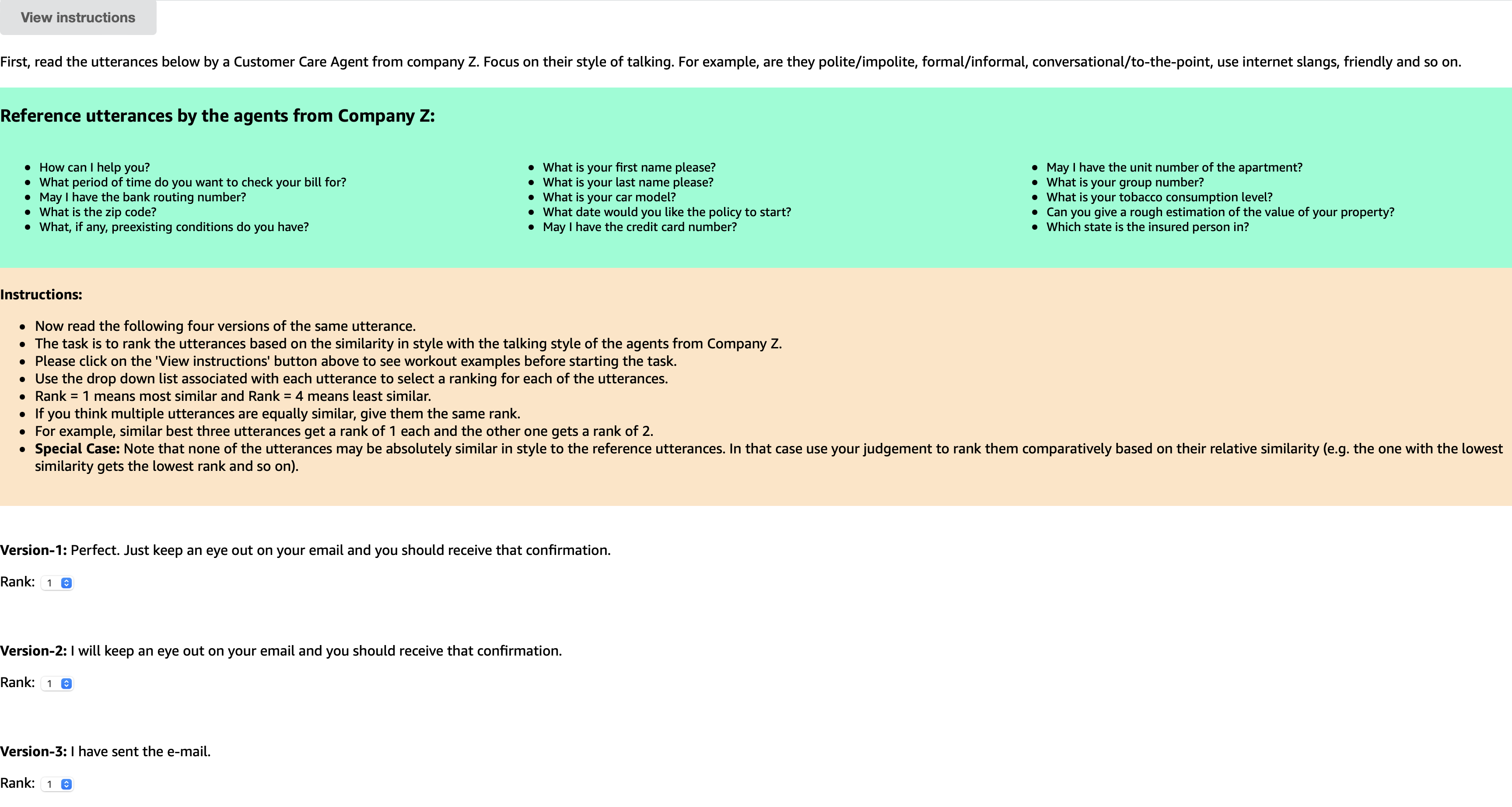}
  \caption{Human evaluation UI for the evaluation of style strength.}
  \label{fig:human-evaluation-prompt-style-strength}
\end{figure*}

\begin{figure*}[h!]
  \centering
  \includegraphics[width=1\textwidth]{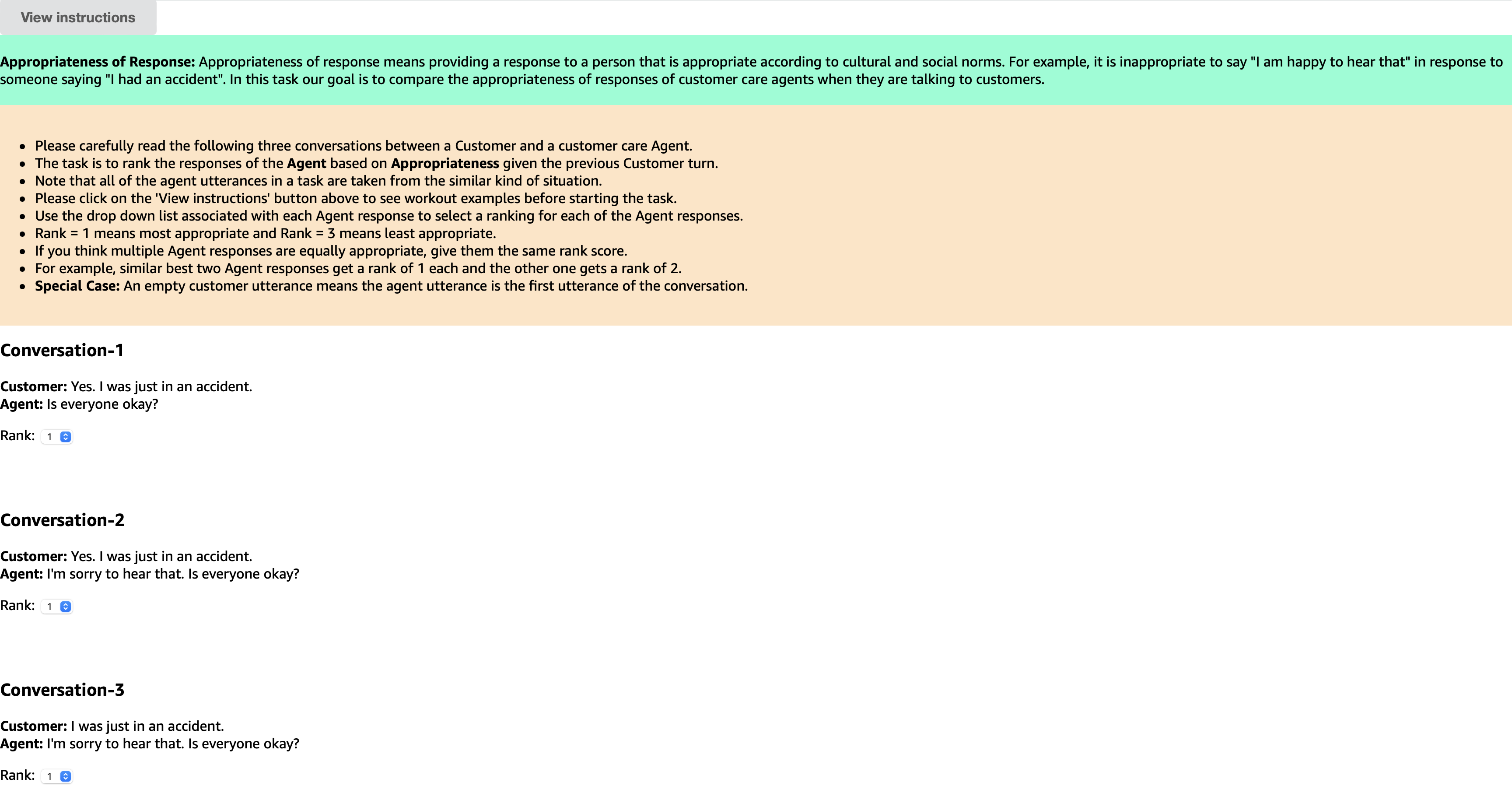}
  \caption{Human evaluation UI for the evaluation of appropriateness.}
  \label{fig:human-evaluation-prompt-appropriateness}
\end{figure*}

\begin{figure*}[h!]
  \centering
  \includegraphics[width=1\textwidth]{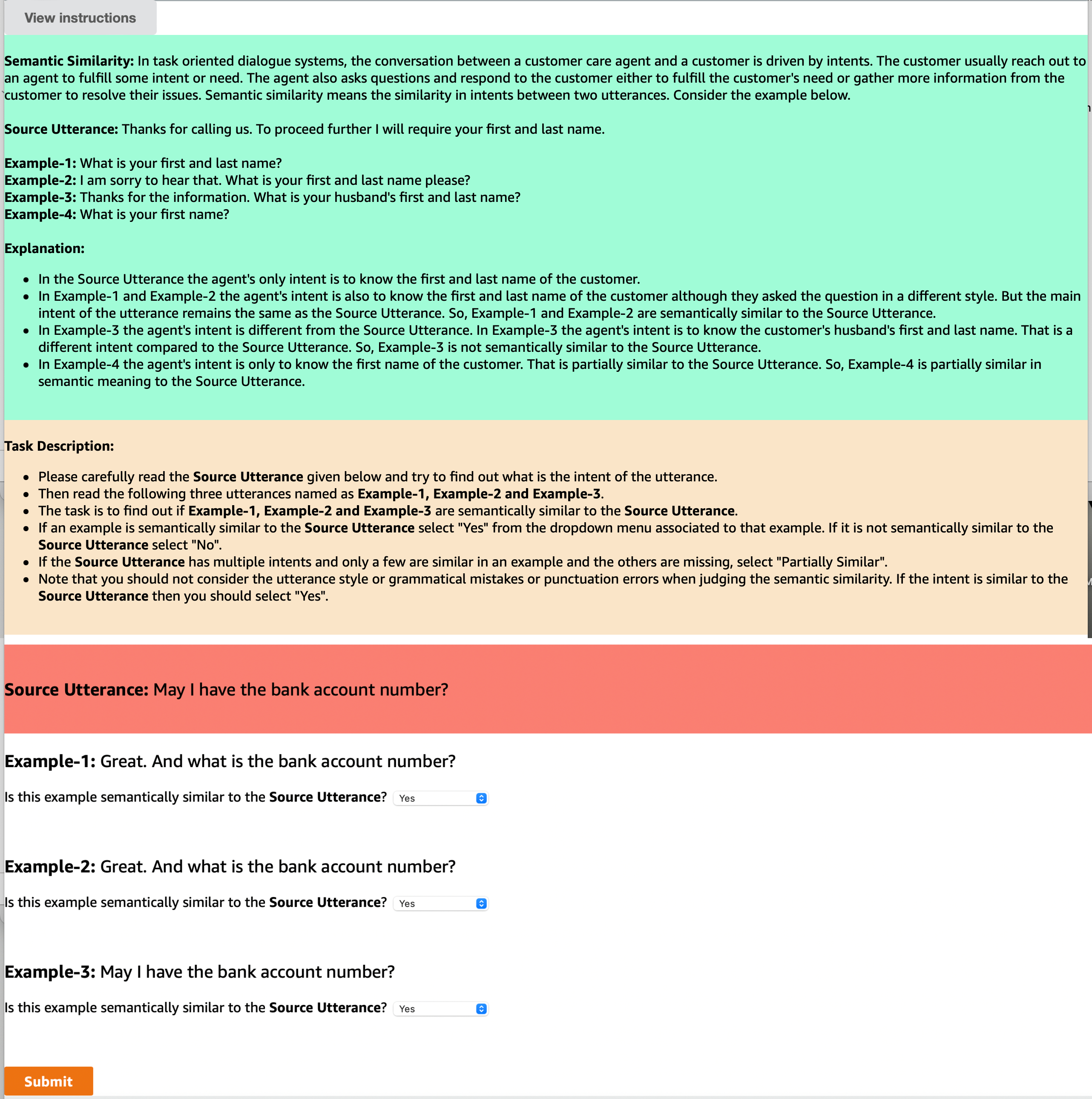}
  \caption{Human evaluation UI for the evaluation of semantic correctness.}
  \label{fig:human-evaluation-prompt-semantic-correctness}
\end{figure*}

Each data point was evaluated by three human evaluators. We worked with professional data linguists who are fluent in English. They were compensated at hourly basis which was in accordance with the standard compensation rate in the United States. They were first trained on the tasks. Specifically, they were briefed on what we mean by style strength, appropriateness, and semantic correctness. Worked-out examples were provided to them. The model names were hidden from the annotators and the four versions were presented in a randomly shuffled order for each example. For ranking-based evaluation in style strength and appropriateness, the human evaluators were instructed to rank the various style-transferred versions from various models based on their style strength and appropriateness. For example, when evaluating among 3 models, a rank of 1 means it has the highest style strength or appropriateness and a rank of 3 means the lowest style strength or appropriateness. The annotators were instructed to provide two style-transferred versions the same rank if they were equal in style strength or appropriateness. For the evaluation of semantic correctness, the human evaluators were presented with the source utterance and the style-transferred versions of the source utterance by each of the models. Then we asked them for each style transferred version if it is semantically similar, partially similar, or dissimilar to the source utterance. Each data point in all of the evaluation metrics is evaluated by three human evaluators. The annotation UIs for style strength, appropriateness, and semantic correctness evaluation tasks are shown in Figures \ref{fig:human-evaluation-prompt-style-strength}, \ref{fig:human-evaluation-prompt-appropriateness}, \ref{fig:human-evaluation-prompt-semantic-correctness}, respectively.

\subsection{Inter-Annotator Agreement} 
\subsubsection{Style Strength and Appropriateness} For measuring the inter-annotator agreement in ranking evaluations for style strength and appropriateness, we use Spearman's Rank Correlation Coefficient \cite{zar2005spearman}. We take the average Spearman's Rank Correlation Coefficient between each pair of human annotators for each data point as an agreement measure. It ranges from -1 to +1 where -1 means absolute disagreement and +1 means absolute agreement.

\subsubsection{Semantic Correctness} For measuring inter-annotator agreement in semantic correctness evaluation task which is categorical, we use Krippendorff’s $\alpha$ \cite{krippendorff2004measuring}. It ranges from -1 to +1 where $\alpha$ = +1 means perfect agreement, and $\alpha$ = -1 means no agreement.

\subsubsection{Agreement Scores} The inter-annotator agreement in all of the tasks are shown in Table \ref{tab:inter-annotator-agreement}. Note that, for calculating agreement in the semantic correctness evaluation task, all of the data points are aggregated to measure the agreement score as they represent categorical evaluation measures. On the other hand, that is not possible in case of ranking based evaluations for style strength and appropriateness. So, we measure the agreement for each data point and take the average agreement over all data points. We can see in the Table \ref{tab:inter-annotator-agreement} that in all of the cases we get strong agreement ($>0.70$) among the annotators for the style strength and appropriateness evaluation. The only exception is the case of style strength evaluation task in the direction of $H_1$ $\rightarrow$ $H_2$, using the GPT-NeoX model. The agreement score is slightly lower ($0.69$) in this case. Our insight is that these two directions are basically human styles and difference between them is very subtle. As a result, it is difficult for humans as well to differentiate among them. This pattern is observed when doing the automatic evaluation as well. 

In case of semantic correctness evaluation task, we always get strong agreement among annotators ($>0.75$).


\subsection{Scaling Ranking Scores} In the style strength and appropriateness evaluation tasks we use ranking based measure among the output from various models. For example, when evaluating among 3 models, a rank of 1 means it has highest style strength or appropriateness and a rank of 3 means the lowest style strength or appropriateness. We scale these rank scores in the range between 0 to 1 where a higher score means higher style strength or appropriateness. The ranking were scaled for each data point using the following formula.

For each data point, if the number of versions to be ranked is $k$ and ranking of a version $i$ $(i \in {1, ..., k})$ is $r_{i}$, then the reverse rank score, $r_{i}^{rev}=k-r_{i}+1$. Now, the scaled rank score, $r_{i}^{scaled}=\frac{r_{i}^{rev}-\min_{j \in 1, ..., k}r_{j}^{rev}}{\max_{j \in 1, ..., k}r_{j}^{rev} - \min_{j \in 1, ..., k}r_{j}^{rev}}$. We average over all human evaluators' scaled ranking score to get the final scaled ranking score for a data point.

\subsection{Pairwise Comparison Among Models}\label{appendix:pairwise-comparison} The pairwise comparison among various versions of the models for style strength and appropriateness are shown in Table \ref{tab:human-evaluation-results-pairwise-comparison}. This table represents the statistics on the percentage of time a model is ranked higher in the style strength and appropriateness evaluation by humans, than the other in a pair-wise manner.

\begin{table*}[t!]
\centering

\resizebox{2\columnwidth}{!}{%
\begin{tabular}
{>{\centering\arraybackslash}m{0.5cm}|>{\arraybackslash}m{1.7cm}>{\centering\arraybackslash}m{1.5cm}>{\centering\arraybackslash}m{1.5cm}|>{\centering\arraybackslash}m{1.5cm}>{\centering\arraybackslash}m{1.5cm}|>{\centering\arraybackslash}m{1.5cm}>{\centering\arraybackslash}m{1.5cm}||>{\centering\arraybackslash}m{1.5cm}>{\centering\arraybackslash}m{1.5cm}|>{\centering\arraybackslash}m{1.5cm}>{\centering\arraybackslash}m{1.5cm}|>{\centering\arraybackslash}m{1.5cm}>{\centering\arraybackslash}m{1.5cm}}
\toprule
    \multicolumn{2}{c}{} & \multicolumn{6}{c}{\textbf{GPT-NeoX (20B)}} &  \multicolumn{6}{c}{\textbf{Bigscience-Bloom (176B)}}\\
    \cmidrule(lr){3-8}\cmidrule(lr){9-14}
    & \textbf{\makecell[c]{Style\\Directions}} & \textbf{$U>C_1$} & \textbf{$U>C_2$} & \textbf{$C_1>U$} & \textbf{$C_1>C_2$} & \textbf{$C_2>U$} & \textbf{$C_2>C_1$} & \textbf{$U>C_1$} & \textbf{$U>C_2$} & \textbf{$C_1>U$} & \textbf{$C_1>C_2$} & \textbf{$C_2>U$} & \textbf{$C_2>C_1$}\\
    \cmidrule(lr){1-2}\cmidrule(lr){3-8}\cmidrule(lr){9-14}
    
    \multirow{4}{*}{\textbf{\rotatebox[origin=c]{90}{Style Strength}}}
    & $H_1 \rightarrow B$   & 31.0 & 48.7 & 11.5 & 30.1 & 12.4 & 9.7 & 25.9 & 37.1 & 5.2 & 26.7 & 16.4 & 21.6\\
    & $H_1 \rightarrow H_2$ & 24.3 & 31.4 & 24.3 & 19.3 & 25.7 & 16.4 & 15.9 & 67.3 & 31.0 & 69.9 & 13.3 & 8.8\\
    & $B \rightarrow H_1$   & 14.7 & 44.1 & 13.7 & 44.1 & 17.6 & 20.6 & 35.0 & 26.0 & 17.9 & 21.1 & 46.3 & 48.8\\
    & $B \rightarrow H_2$   & 62.7 & 67.9 & 1.5 & 28.4 & 3.0 & 3.7 & 34.4 & 78.4 & 17.6 & 74.4 & 2.4 & 3.2\\
    \cmidrule(lr){2-8}\cmidrule(lr){9-14} 
    & Average               & 33.2 & 48.0 & 12.8 & 30.5 & 14.7 & 12.6 & 27.8 & 52.2 & 17.9 & 48.0 & 19.6 & 20.6\\
    
    \midrule
    
    \multirow{4}{*}{\textbf{\rotatebox[origin=c]{90}{Appropriate.}}}
    & $H_1 \rightarrow B$   & 3.5 & 1.8 & 9.7 & 3.5 & 8.8 & 5.3 & 3.4 & 6.0 & 5.2 & 6.0 & 6.9 & 4.3\\
    & $H_1 \rightarrow H_2$ & 2.1 & 3.6 & 27.1 & 4.3 & 29.3 & 3.6 & 6.2 & 5.3 & 13.3 & 5.3 & 14.2 & 7.1\\
    & $B \rightarrow H_1$   & 1.0 & 2.0 & 0.0 & 2.0 & 0.0 & 1.0 & 4.1 & 7.3 & 0.0 & 6.5 & 1.6 & 4.1\\
    & $B \rightarrow H_2$   & 0.0 & 3.7 & 60.4 & 3.7 & 59.7 & 0.0 & 4.0 & 5.6 & 8.8 & 5.6 & 9.6 & 4.8\\
    \cmidrule(lr){2-8}\cmidrule(lr){9-14} 
    & Average               & 1.7 & 2.8 & 24.3 & 3.4 & 24.5 & 2.5 & 4.4 & 6.1 & 6.8 & 5.9 & 8.1 & 5.1\\
    
    \bottomrule
\end{tabular}}

\caption{Human evaluation results on style strength and appropriateness. The table presents a pair-wise comparison among three versions of our model - utterance level style transfer (denoted as $U$), 2-turns conversation level style transfer (denoted as $C_1$), and 4/5-turns conversation level style transfer (denoted as $C_2$). Each cell represents the $\%$ of time a model is ranked higher than the other by the human evaluators. For example, column $U>C_1$ represents the $\%$ of time the utterance level style transfer model is ranked higher than the 2-turns conversation level style transfer.}

\label{tab:human-evaluation-results-pairwise-comparison}
\end{table*}

\section{Downstream Application: Intent Classification}\label{appendix:downstream-application}

\subsection{Dataset}
We take the three domains of DSTC11 dataset namely, Insurance, Banking and Finance for this task. In this dataset, mostly the customer utterances are annotated for intents. We take the human-to-human conversations as training data and human-to-bot conversations as test data. We consider intent classes having at least 20 training utterances for this study. Then we randomly select $90\%$ of training data from each intent class as training set and select rest of the $10\%$ as validation set. The training, test and validation data statistics for each of the domains are shown in Table \ref{tab:ic-data-statistics}.


\begin{table}[h!]
\centering
\resizebox{1\columnwidth}{!}{%
\begin{tabular}
{>{\arraybackslash}m{3.2cm}>{\centering\arraybackslash}m{1.9cm}>{\centering\arraybackslash}m{1.7cm}>{\centering\arraybackslash}m{1.8cm}}

\toprule
    & \textbf{Insurance (21 classes)} & \textbf{Banking (9 classes)} & \textbf{Finance (23 classes)}\\
    \cmidrule(lr){2-4}
    \# of train. utterances & 849   & 1095  & 1169\\
    \# of valid. utterances & 106   & 124   & 142\\
    \# of test utterances   & 653   & 144   & 756\\
    \bottomrule
\end{tabular}}
\caption{Intent classification dataset statistics.}\label{tab:ic-data-statistics}
\end{table}

\subsection{Few-Shot Style Transfer of Training Data}
In this dataset, mostly customer utterances are annotated for intent classes. So, we perform few-shot style transfer of the customer utterances only, using the same procedure that we followed for agent utterances style transfer. We found out that customers are more conversational when talking to a human agent compared to when talking to a chatbot agent. So, we use few-shot customer utterances from the human-to-bot conversations to transfer the style of customers in human-to-human conversations. Then use this style transferred data for training an intent classifier. We use a 10-shot setting with dynamic prompt selection based on semantic similarity. 


\subsection{Intent Classification Results}
We compare the performance of the intent classifier when trained on human-to-human conversations vs. training on human-to-human conversations that are transferred to human-to-bot style. We ran an ablation where we experimented with utterance level style transfer and 2-turns conversation level style transfer as these two methods yielded better style strength in our studies. We ran this ablation using only banking and finance domains out of the thee domains. The classification was done 10 times with 10 random seeds for each domain. A RoBERTa-based \cite{liu2019roberta} text classifier was used to perform the intent classification task. We encoded each utterance using RoBERTa where the embedding of the [CLS] token of the last layer was used as a representation of the utterance. This representation was used for intent classification. The average classification results are shown in Table \ref{tab:ic-results-details}. Overall, the utterance level style transfer yields the best intent classification results as it achieves the best style strength of the test domain (human-to-bot style).

%

\begin{table*}[h!]
\centering
\resizebox{2\columnwidth}{!}{%
\begin{tabular}
{>{\arraybackslash}m{5cm}>{\centering\arraybackslash}m{1.8cm}>{\centering\arraybackslash}m{1.8cm}>{\centering\arraybackslash}m{2.1cm}>{\centering\arraybackslash}m{1.8cm}>{\centering\arraybackslash}m{1.8cm}>{\centering\arraybackslash}m{2.1cm}>{\centering\arraybackslash}m{1.8cm}>{\centering\arraybackslash}m{1.8cm}>{\centering\arraybackslash}m{2.1cm}}

\toprule
    \multicolumn{1}{c}{} & \multicolumn{9}{c}{\textbf{F1 score on test data (human-to-bot conversations)}}\\
    \cmidrule(lr){2-10}
    \textbf{Training data} & \multicolumn{3}{c}{\textbf{Insurance (21 classes)}} & \multicolumn{3}{c}{\textbf{Banking (9 classes)}} & \multicolumn{3}{c}{\textbf{Finance (23 classes)}}\\
    & \textbf{Macro F1} & \textbf{Micro F1} & \textbf{Weighted F1} & \textbf{Macro F1} & \textbf{Micro F1} & \textbf{Weighted F1} & \textbf{Macro F1} & \textbf{Micro F1} & \textbf{Weighted F1}\\
    \cmidrule(lr){1-1}\cmidrule(lr){2-4}\cmidrule(lr){5-7}\cmidrule(lr){8-10}
    hum.-to-hum. conv. & 92.39 (0.5) & 92.96 (0.4) & 92.46 (0.5) & 94.43 (2.1) & 94.44 (2.0) & 94.43 (2.1) & 89.70 (0.6) & \textbf{91.23 (0.6)} & \textbf{90.49 (0.5)}\\
    \hline
    hum.-to-hum. conv. transferred to hum.-to-bot style using 2-turns conversation level style transfer & - & - & - & 94.70 (1.7) & 94.80 (1.7) & 94.70 (1.7) & 89.60 (0.8) & 91.20 (0.6) & 90.40 (0.6)\\
    \hline
    hum.-to-hum. conv. transferred to hum.-to-bot style using utterance level style transfer & \textbf{92.96 (0.5)} & \textbf{93.51 (0.5)} & \textbf{93.00 (0.5)} & \textbf{97.70 (1.3)} & \textbf{97.71 (1.2)} & \textbf{97.70 (1.3)} & \textbf{89.92 (0.5)} & 91.08 (0.4) & 90.34 (0.4)\\
    \bottomrule
\end{tabular}}
\caption{Detailed Intent classification results. The ablation between two types of models - utterance level style transfer and 2-turns conversation level style transfer was performed on two domains - banking and finance. Overall, utterance level style transfer yields the best intent classification F1 scores as it achieves the highest style strength score as the test domain (human-to-bot).}\label{tab:ic-results-details}
\end{table*}

\section{LLM Hyperparameters and Infrastructure Used}\label{appendix:llm-hyperparameters}
We use top-k sampling with temperature, t \cite{holtzman2019curious} as a decoding method for the large language models. t = 0.1 was set for all of our experiments. We ran all of the experiments using PyTorch. Both Bloom and GPT-NeoX were run on a computation node with 8 A100 GPUs.  



\section{Style Transfer Evaluation Results}\label{appendix:style-transfer-eval-results-with-sd}
Table \ref{tab:human-evaluation-results-with-sd} and Table \ref{tab:automatic-evaluation-results-with-sd} presents human and automatic evaluation results for various evaluation metrics with standard deviations.

\begin{table*}[t!]
\centering

\resizebox{2.1\columnwidth}{!}{%
\begin{tabular}
{>{\centering\arraybackslash}m{0.5cm}|>{\arraybackslash}m{1.6cm}>{\centering\arraybackslash}m{2.6cm}>{\centering\arraybackslash}m{3cm}>{\centering\arraybackslash}m{3cm}>{\centering\arraybackslash}m{3cm}|>{\centering\arraybackslash}m{2.6cm}>{\centering\arraybackslash}m{3cm}>{\centering\arraybackslash}m{3cm}>{\centering\arraybackslash}m{3.1cm}}
\toprule
    \multicolumn{2}{c}{} & \multicolumn{4}{c}{\textbf{GPT-NeoX (20B)}} &  \multicolumn{4}{c}{\textbf{Bigscience-Bloom (176B)}} \\
    \cmidrule(lr){3-6}\cmidrule(lr){7-10}
   & \textbf{Style} & \textbf{Original} & \textbf{Utterance Level} & \multicolumn{2}{c}{\textbf{\underline{\hspace{1.5em}Conversation Level Style Transfer\hspace{1.5em}}}} & \textbf{Original} & \textbf{Utterance Level} & \multicolumn{2}{c}{\textbf{\underline{\hspace{1.5em}Conversation Level Style Transfer\hspace{1.5em}}}}\\
   & \textbf{Directions} & \textbf{Utterances} & \textbf{Style Transfer} & \textbf{2-turns convo.} & \textbf{4/5-turns convo.} & \textbf{Utterances} & \textbf{Style Transfer} & \textbf{2-turns convo.} & \textbf{4/5-turns convo.}\\
   \cmidrule(lr){1-2}\cmidrule(lr){3-6}\cmidrule(lr){7-10}
   & & \textbf{Avg. rank score} & \textbf{Avg. rank score} & \textbf{Avg. rank score} & \textbf{Avg. rank score} & \textbf{Avg. rank score} & \textbf{Avg. rank score} & \textbf{Avg. rank score} & \textbf{Avg. rank score}\\
   \cmidrule(lr){3-6}\cmidrule(lr){7-10}
    
    \multirow{4}{*}{\textbf{\rotatebox[origin=c]{90}{Style Strength}}}
    & $H_1 \rightarrow B$ & 0.392 (0.483) & 0.864 (0.336) & 0.714 (0.445) & 0.561 (0.490) & 0.435 (0.482) & 0.876 (0.313) & 0.719 (0.433) & 0.720 (0.446)\\
      & $H_1 \rightarrow H_2$ & 0.15 (0.357) & 0.854 (0.275) & 0.855 (0.267) & 0.838 (0.294) & 0.125 (0.301) & 0.895 (0.227) & 0.924 (0.207) & 0.538 (0.451)\\
      & $B \rightarrow H_1$ & 0.574 (0.489) & 0.851 (0.341) & 0.846 (0.356) & 0.690 (0.452) & 0.378 (0.480) & 0.692 (0.450) & 0.622 (0.472) & 0.856 (0.329)\\
      & $B \rightarrow H_2$ & 0.043 (0.203) & 0.989 (0.073) & 0.805 (0.246) & 0.690 (0.352) & 0.024 (0.111) & 0.958 (0.135) & 0.897 (0.219) & 0.484 (0.424)\\
    \cmidrule(lr){2-6}\cmidrule(lr){7-10} 
    & Average & 0.290 & \textbf{0.890} & 0.805 & 0.695 & 0.241 & \textbf{0.855} & 0.791 & 0.650\\
    \cmidrule(lr){1-10}
    
    
    \multirow{4}{*}{\textbf{\rotatebox[origin=c]{90}{Appropriate.}}}
    & $H_1 \rightarrow B$ &  0.997 (0.054) & 0.943 (0.231) & 0.971 (0.169) & 0.979 (0.142) & 0.991 (0.092) & 0.968 (0.175) & 0.974 (0.159) & 0.966 (0.183)\\
      & $H_1 \rightarrow H_2$ & 0.980 (0.139) & 0.798 (0.402) & 0.985 (0.121) & 0.977 (0.147) & 0.997 (0.054) & 0.917 (0.275) & 0.972 (0.163) & 0.974 (0.161)\\
      & $B \rightarrow H_1$ & 0.997 (0.057) & 1.0 (0.0) & 0.997 (0.057) & 0.987 (0.114) & 0.995 (0.073) & 0.995 (0.073) & 0.980 (0.139) & 0.968 (0.177)\\
      & $B \rightarrow H_2$ & 0.990 (0.099) & 0.481 (0.500) & 1.00 (0) & 0.978 (0.148) & 0.995 (0.073) & 0.923 (0.267) & 0.957 (0.202) & 0.976 (0.153)\\
    \cmidrule(lr){2-6}\cmidrule(lr){7-10} 
    & Average & 0.991 & 0.806 & \textbf{0.988} & 0.980 & 0.995 & 0.951 & \textbf{0.971} & \textbf{0.971}\\
    \cmidrule(lr){1-10}

    
    \multirow{5}{*}{\textbf{\rotatebox[origin=c]{90}{Semantic Correct.}}}
    & & & \textbf{yes-partially-no} & \textbf{yes-partially-no} & \textbf{yes-partially-no} & & \textbf{yes-partially-no} & \textbf{yes-partially-no} & \textbf{yes-partially-no}\\
    \cmidrule(lr){4-6}\cmidrule(lr){8-10}
    & $H_1 \rightarrow B$ &  & 0.885-0.026-0.089 & 0.938-0.009-0.053 & 0.920-0.027-0.053 &  & 0.948-0-0.052 & 0.974-0-0.026 & 0.767-0.035-0.198\\
    & $H_1 \rightarrow H_2$ &  & 0.921-0.007-0.071 & 0.964-0.007-0.029 & 0.943-0.021-0.036 &  & 0.894-0.009-0.097 & 0.956-0-0.044 & 0.841-0.009-0.150\\
    & $B \rightarrow H_1$ &  & 1-0-0 & 0.980-0-0.020 & 0.961-0.019-0.020 &  & 1-0-0 & 0.968-0-0.032 & 0.862-0-0.138\\
    & $B \rightarrow H_2$ &  & 0.993-0-0.007 & 1-0-0 & 1-0-0 &  & 1-0-0 & 0.992-0-0.008 & 0.880-0-0.120\\
    \cmidrule(lr){2-6}\cmidrule(lr){7-10} 
    & Average & & 0.95-0.008-0.042 & \textbf{0.97-0.004-0.026} & 0.956-0.017-0.027 & & 0.961-0.002-0.037 & \textbf{0.973-0-0.027} & 0.838-0.011-0.151\\
    \bottomrule
\end{tabular}}
\caption{Human evaluation results for utterance level (baseline) and conversation level style transfer with GPT-NeoX and Bigscience-Bloom LLMs using our model. The best average score over all style dimensions are marked in bold. Utterance level style transfer achieves higher style strength but conversation level style transfers yield more appropriate and semantically correct responses.}

\label{tab:human-evaluation-results-with-sd}
\end{table*}

\begin{table*}[t!]
\centering

\resizebox{2.1\columnwidth}{!}{%
\begin{tabular}
{>{\centering\arraybackslash}m{0.5cm}|>{\arraybackslash}m{1.6cm}>{\centering\arraybackslash}m{2.2cm}>{\centering\arraybackslash}m{3cm}>{\centering\arraybackslash}m{3cm}>{\centering\arraybackslash}m{3cm}|>{\centering\arraybackslash}m{2.2cm}>{\centering\arraybackslash}m{3cm}>{\centering\arraybackslash}m{3cm}>{\centering\arraybackslash}m{3.1cm}}
\toprule
    \multicolumn{2}{c}{} & \multicolumn{4}{c}{\textbf{GPT-NeoX (20B)}} &  \multicolumn{4}{c}{\textbf{Bigscience-Bloom (176B)}} \\
    \cmidrule(lr){3-6}\cmidrule(lr){7-10}
   & \textbf{Style} & \textbf{Original} & \textbf{Utterance Level} & \multicolumn{2}{c}{\textbf{\underline{\hspace{1.5em}Conversation Level Style Transfer\hspace{1.5em}}}} & \textbf{Original} & \textbf{Utterance Level} & \multicolumn{2}{c}{\textbf{\underline{\hspace{1.5em}Conversation Level Style Transfer\hspace{1.5em}}}}\\
   & \textbf{Directions} & \textbf{Utterances} & \textbf{Style Transfer} & \textbf{2-turns convo.} & \textbf{4/5-turns convo.} & \textbf{Utterances} & \textbf{Style Transfer} & \textbf{2-turns convo.} & \textbf{4/5-turns convo.}\\
   \cmidrule(lr){1-2}\cmidrule(lr){3-6}\cmidrule(lr){7-10}
   & & \textbf{Avg. target style strength} & \textbf{Avg. target style strength} & \textbf{Avg. target style strength} & \textbf{Avg. target style strength} & \textbf{Avg. target style strength} & \textbf{Avg. target style strength} & \textbf{Avg. target style strength} & \textbf{Avg. target style strength}\\
   \cmidrule(lr){3-6}\cmidrule(lr){7-10}
    \multirow{4}{*}{\textbf{\rotatebox[origin=c]{90}{Style Strength}}}
    & $H_1 \rightarrow B$
    & 0.038 (0.184) & 0.224 (0.406) & 0.184 (0.373) & 0.154 (0.358) & 0.036 (0.181) & 0.209 (0.400) & 0.196 (0.388) & 0.256 (0.427) \\
    & $H_1 \rightarrow H_2$ & 0.129 (0.282) & 0.215 (0.349) & 0.192 (0.340) & 0.161 (0.308) & 0.139 (0.200) & 0.246 (0.370) & 0.236 (0.381) & 0.169 (0.308) \\
    & $B \rightarrow H_1$    & 0.001 (0.001) & 0.500 (0.493) & 0.388 (0.485) & 0.174 (0.377) & 0.001 (0.001) & 0.589 (0.485) & 0.463 (0.496) & 0.782 (0.410) \\
    & $B \rightarrow H_2$    & 0             & 0.589 (0.464) & 0.131 (0.324) & 0             & 0             & 0.286 (0.386) & 0.192 (0.342) & 0.126 (0.328) \\
    \cmidrule(lr){2-6}\cmidrule(lr){7-10} 
    & Average & 0.042 & \textbf{0.382} & 0.224 & 0.122 & 0.044 & \textbf{0.333} & 0.272 & \textbf{0.333} \\
    \cmidrule(lr){1-10}
     
    \multirow{5}{*}{\textbf{\rotatebox[origin=c]{90}{Semantic Correct.}}}
    & & & \textbf{Avg. semantic sim. to original utt.} & \textbf{Avg. semantic sim. to original utt.} & \textbf{Avg. semantic sim. to original utt.} & & \textbf{Avg. semantic sim. to original utt.} & \textbf{Avg. semantic sim. to original utt.} & \textbf{Avg. semantic sim. to original utt.}\\
    \cmidrule(lr){4-6}\cmidrule(lr){8-10}
    & $H_1 \rightarrow B$    & & 0.827 (0.358) & 0.875 (0.211) & 0.912 (0.184) & & 0.795 (0.207) & 0.855 (0.203) & 0.778 (0.239)\\
    & $H_1 \rightarrow H_2$ & & 0.774 (0.222) & 0.842 (0.196) & 0.875 (0.174) & & 0.752 (0.231) & 0.802 (0.201) & 0.793 (0.225)\\
    & $B \rightarrow H_1$    & & 0.860 (0.184) & 0.844 (0.195) & 0.919 (0.128) & & 0.863 (0.162) & 0.850 (0.187) & 0.787 (0.201)\\
    & $B \rightarrow H_2$    & & 0.749 (0.159) & 0.801 (0.097) & 0.828 (0.106) & & 0.756 (0.120) & 0.748 (0.116) & 0.781 (0.164)\\
    
    \cmidrule(lr){2-6}\cmidrule(lr){7-10} 
    & Average & & 0.803 & 0.841 & \textbf{0.884} & & 0.792 & \textbf{0.814} & 0.785 \\
    \bottomrule
\end{tabular}}

\caption{Automatic style strength and semantic correctness evaluation results for utterance level (baseline) and conversation level style transfer with GPT-NeoX and Bigscience-Bloom LLMs using our model. Utterance level style transfer achieves higher style strength and conversation level style transfers yield more semantically similar responses.}


\label{tab:automatic-evaluation-results-with-sd}
\end{table*}

\section{Qualitative Examples}\label{appendix:qualitative-examples}
\begin{figure*}[h]
    \begin{center}
    \begin{subfigure}[t]{1\textwidth}
        \centering
        \includegraphics[width=1\textwidth]{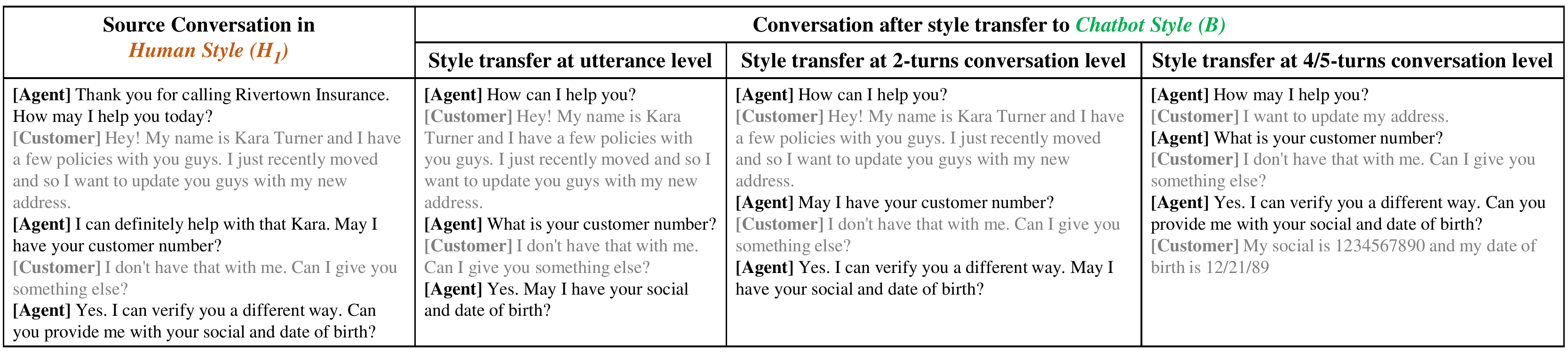}
        \caption{Style transfer from human style ($H_1$) to chatbot style ($B$).}
    \end{subfigure}
    \\
    \begin{subfigure}[t]{1\textwidth}
        \centering
        \includegraphics[width=1\textwidth]{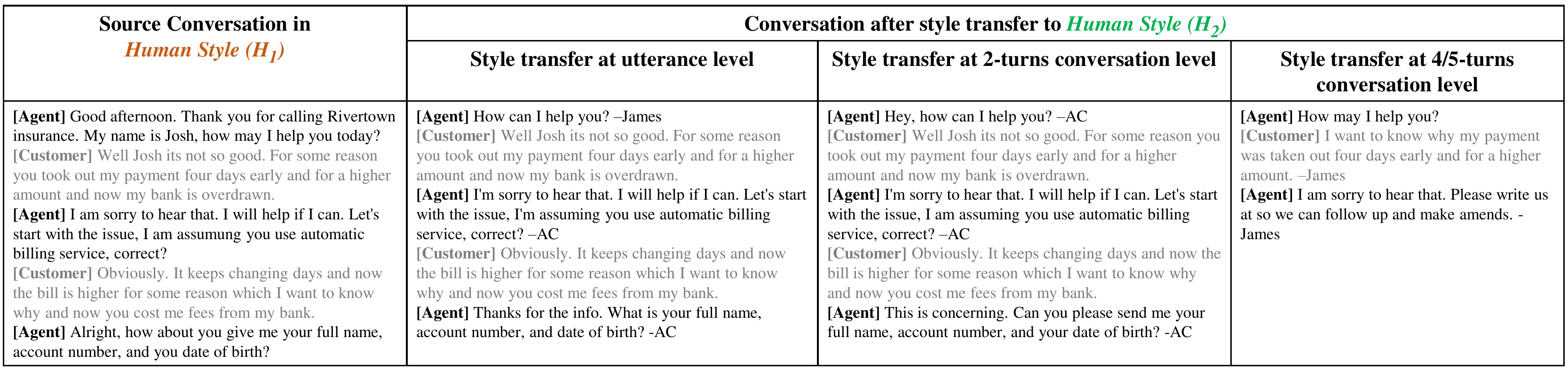}
        \caption{Style transfer from one human style ($H_1$) to another human style ($H_2$).}
    \end{subfigure}
    \\
    \begin{subfigure}[t]{1\textwidth}
        \centering
        \includegraphics[width=1\textwidth]{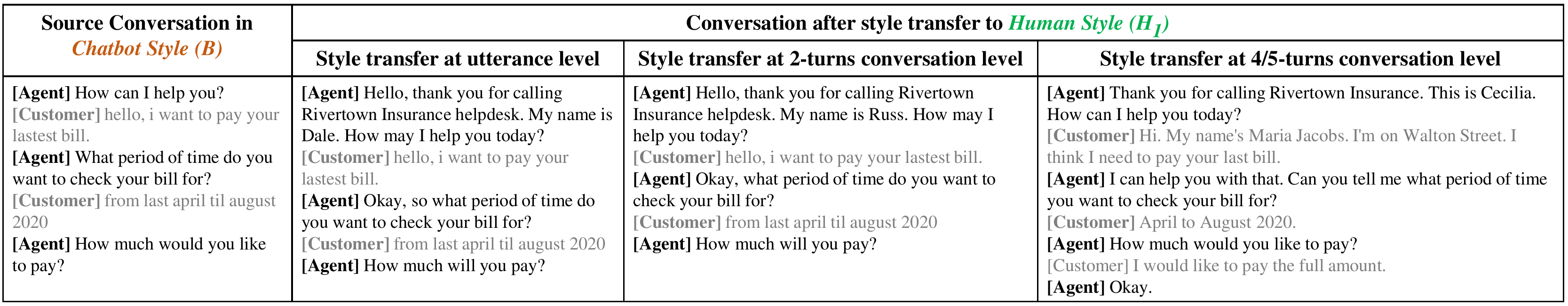}
        \caption{Style transfer from one chatbot style ($B$) to human style ($H_1$).}
    \end{subfigure}
    \\
    \begin{subfigure}[t]{1\textwidth}
        \centering
        \includegraphics[width=1\textwidth]{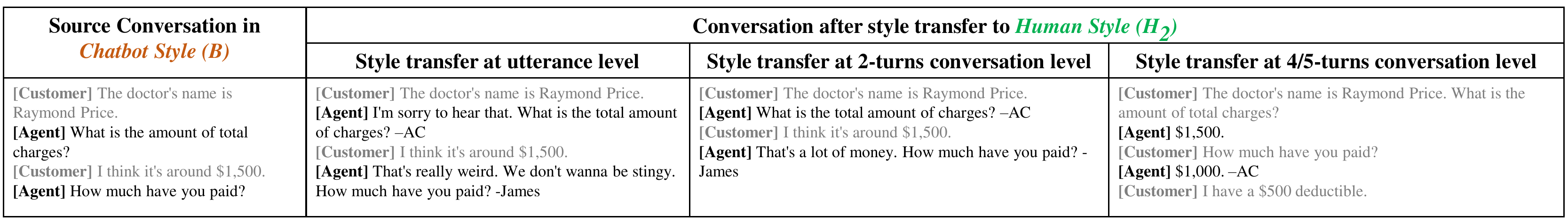}
        \caption{Style transfer from one chatbot style ($B$) to human style ($H_2$).}
    \end{subfigure}
    \caption{Style transfer examples in all directions by various versions of our model. Here, $H_1, H_2, B$ refer to human style from DSTC11 dataset, style of Chipotle agents (from TWCS dataset), and Chatbot style from DSTC11 dataset, respectively.}
    \label{fig:qualitative-examples-all-directions}
    \end{center}
\end{figure*}

Figure \ref{fig:qualitative-examples-all-directions} presents style transfer examples in all directions by various versions of our model.

\section{Error Analysis for Bigscience-Bloom}\label{appendix:error-analysis-bigscience-bloom}
Examples of some common types of errors observed in 4/5-turns conversation level style transfer using Bigscience-Bloom are shown in Table \ref{tab:error-analysis-bigscience-bloom}.

\begin{table*}
\centering
\resizebox{2\columnwidth}{!}{%
\begin{tabular}
{>{\centering\arraybackslash}m{10cm}|>{\centering\arraybackslash}m{10cm}}

\toprule

\textbf{Original 4/5-Turns Conversations} & \textbf{Style Transferred Version by Bigscience-Bloom}\\
\cmidrule(lr){1-1}\cmidrule(lr){2-2}
\textbf{Source Style: $B$} & \textbf{Target Style: $H_2$}\\
\cmidrule(lr){1-1}\cmidrule(lr){2-2}
\makecell[l]{\textbf{[Customer]} I would like to pay \$500\\\textbf{[Agent]} Is this a checking or saving account?\\\textbf{[Customer]} This is a checking account\\\textbf{[Agent]} May I have the bank account number?}

& \makecell[l]{\textbf{[Customer]} \$500. Is this a checking or saving account?\\\textbf{[Agent]} Checking account.\\\textbf{[Customer]} What is the bank account number?\\\textbf{[Agent]} 998201450. -James}\\

\cmidrule(lr){1-2}
\textbf{Source Style: $H_1$} & \textbf{Target Style: $B$}\\
\cmidrule(lr){1-1}\cmidrule(lr){2-2}
\makecell[l]{\textbf{[Agent]} Thank you for calling Rivertown Insurance. How may\\I help you today?\\\textbf{[Customer]} Yeah I'm trying to register my car and they need\\my insurance number and I don't have it.\\\textbf{[Agent]} That's not a problem. I can look it up for you!\\\textbf{[Customer]} Great, thanks. Trying to register this car has been\\a nightmare. Hopefully this will be easier.\\\textbf{[Agent]} I understand. When I registered mine it took almost\\a week. It was crazy.}

& \makecell[l]{\textbf{[Agent]} How may I help you?\\\textbf{[Customer]} I need to register my car.\\\textbf{[Agent]} May I have your policy number?\\\textbf{[Customer]} I don't have it.}\\

\bottomrule
\end{tabular}}
\caption{Error analysis of Bigscience-Bloom when performing 4/5-turns conversation level style transfer. Sometimes customer and agent turns are swapped as shown in the upper conversation and sometimes unrelated agent responses are generated as shown in the lower conversation. Here, $H_1, H_2, B$ refer to human style from DSTC11 dataset, style of Chipotle agents (from TWCS dataset), and Chatbot style from DSTC11 dataset, respectively.}
\label{tab:error-analysis-bigscience-bloom}
\end{table*}

\section{Style Discriminator Models}\label{appendix:style-discriminator}

We train RoBERTa-based binary text classifiers to classify between the source and the target styles. Training data for these classifiers are obtained from the residual data after selecting the test and the validation sets as described in Table \ref{tab:dataset-summary}. We treat the confidence scores of these classifiers as the style strength scores. We balance the training data for both of the classes when training these classifiers. For training the classifiers to differentiate between styles ($H_{1}$, $B$), ($H_{1}$, $H_{2}$), ($H_{2}$, $B$), we randomly sampled $4,875$, $1,792$ and $1,792$ agent utterances from each class, respectively. $10\%$ of the data were held out as a validation set. We encoded each agent utterance using a RoBERTa model where the embedding of the [CLS] token of the last layer was used as a representation of the utterance. We used this representation for the classification of the style domain. We stopped training when the validation accuracy did not improve for consecutive two epochs. The validation accuracy of the classifiers to differentiate between styles ($H_{1}$, $B$), ($H_{1}$, $H_{2}$), ($H_{2}$, $B$) were $99.89\%$, $93.3\%$ and $100\%$, respectively. Note that, style $H_{2}$ has a unique property that each agent signs their name after their responses preceded by a hyphen. If we train a classifier to identify style $H_{2}$, it always yielded an accuracy of $100\%$ because of the specific signature format. As a result, other stylistic properties such as vocabulary usage, crispness, conversational, and so on were missed out by the style classifier. Hence, for training the classifiers involving this style class, we removed these signatures as a preprocessing step.

\section{Effect on Observed Style Properties after Style Transfer}\label{appendix:effect-on-observed-properties}
In this section, we examine the effect of style transfer on the observed style properties as described in Table \ref{tab:quantitative-style}. Note that, our main observation in this paper is that conversation styles are difficult to determine and characterize using a fixed set of attributes (as described in Sections \ref{sec:problem-formulation} and \ref{sec:experiment}). However, we examine the effect of style transfer on the observed properties in Table \ref{tab:quantitative-style} for the sake of completeness of our experiments and sanity checking of our models' performances. As described in Section \ref{sec:experiment}, conversation styles are rather holistic and the true style of the domains $H_1$, $H_2$, and $B$ go beyond these observed properties and they are difficult to characterize using a fixed set of attributes.

We present the effect of style transfer on crispness, diversity in vocabulary, and the structural attribute - signing names at the end of responses in Table \ref{tab:appendix-observed-style-properties}. We have observed in Table \ref{tab:quantitative-style} that chatbot agent ($B$) responses are crisper than human agent responses ($H_1$, $H_2$). It can be observed in Table \ref{tab:appendix-observed-style-properties} that when transferring from human style ($H_1$) to chatbot style ($B$), the average number of words per agent turn is decreased by all of the models to make them crisp. Conversely, the number of average words per agent turn is increased by most of the models when transferring from chatbot style ($B$) to human styles ($H_1$, $H_2$) to make them more conversational. 

We also observed in Table \ref{tab:quantitative-style} that human agents use diverse vocabulary compared to chatbot agents. Consequently, we observe in Table \ref{tab:appendix-observed-style-properties} that vocabulary is made less diverse (compressed) when transferring from human ($H_1$) to chatbot style ($B$) and more diverse (expanded) when transferring from chatbot ($B$) to human styles ($H_1$, $H_2$).

Signing names at the end of a response is a unique structural style property of the style $H_2$ (Table \ref{tab:quantitative-style}), hence, this style property is obtained by the models only when transferring a source style to $H_2$. We can observe in Table \ref{tab:appendix-observed-style-properties} that all models successfully achieve this property when transferring a source style to the style $H_2$ except in the 4/5-conversation-level-style-transfer using Bigscience-Bloom.

This evaluation proves that the proposed models can successfully achieve the observed style properties during style transfer.

\begin{table*}[ht!]
\centering

\resizebox{2\columnwidth}{!}{%
\begin{tabular}
{>{\arraybackslash}m{2.5cm}>{\arraybackslash}m{1.6cm}>{\centering\arraybackslash}m{2cm}|>{\centering\arraybackslash}m{2cm}>{\centering\arraybackslash}m{2cm}>{\centering\arraybackslash}m{2cm}|>{\centering\arraybackslash}m{2cm}>{\centering\arraybackslash}m{2cm}>{\centering\arraybackslash}m{2cm}}
\toprule
    \multicolumn{1}{c}{} & \multicolumn{1}{c}{} & \multicolumn{1}{c}{} & \multicolumn{6}{c}{\textbf{Value after style transfer}} \\
    \cmidrule(lr){4-9}
    &  &  & \multicolumn{3}{c}{\textbf{GPT-NeoX (20B)}} & \multicolumn{3}{c}{\textbf{Bigscience-Bloom (176B)}}\\
    \cmidrule(lr){4-6}\cmidrule(lr){7-9}
    \textbf{\makecell[c]{Observed\\properties}} & \textbf{\makecell[c]{Style\\directions}} & \textbf{Value before style transfer} & \textbf{Utterance Level} & \textbf{2 Turns Conv. Level} & \textbf{4/5 Turns Conv. Level} & \textbf{Utterance Level} & \textbf{2 Turns Conv. Level} & \textbf{4/5 Turns Conv. Level}\\
    \cmidrule(lr){1-9}

    \multirow{4}{*}{\textbf{\makecell[l]{Avg. \# of words\\per agent turn\\(Crispness)}}}
    & $H_1 \rightarrow B$    & $12.76\pm8.22$ & $10.5\pm8.43$ & $11.47\pm8.56$ & $10.94\pm8.43$ & $9.12\pm7.25$ & $10.92\pm8.35$ & $8.63\pm6.35$\\
    & $H_1 \rightarrow H_2$   & $12.76\pm8.22$ & $11.9\pm7.43$ & $12.05\pm7.73$ & $11.5\pm7.55$ & $11.02\pm7.69$ & $11.87\pm7.41$ & $9.22\pm6.29$\\
    & $B \rightarrow H_1$   & $6.97\pm1.85$ & $8.08\pm3.96$ & $8.39\pm3.94$ & $7.77\pm3.86$ & $9.14\pm4.66$ & $8.68\pm4.11$ & $9.53\pm7.07$\\
    & $B \rightarrow H_2$   & $6.97\pm1.85$ & $9.18\pm2.66$ & $6.71\pm1.95$ & $6.53\pm2.31$ & $8.75\pm2.6$ & $7.6\pm2.72$ & $5.69\pm2.84$\\
    \cmidrule(lr){1-9}
    
    \multirow{4}{*}{\textbf{\makecell[l]{Vocabulary size\\(Diversity)}}}
    & $H_1 \rightarrow B$   & $527$ & $477$ & $489$ & $458$ & $424$ & $488$ & $371$\\   
    & $H_1 \rightarrow H_2$ & $527$ & $463$ & $492$ & $463$ & $437$ & $473$ & $383$\\
    & $B \rightarrow H_1$   & $97$ & $106$ & $126$ & $149$ & $125$ & $139$ & $186$\\
    & $B \rightarrow H_2$   & $97$ & $125$ & $103$ & $119$ & $117$ & $145$ & $134$\\
    \cmidrule(lr){1-9}
    
    \multirow{4}{*}{\textbf{\makecell[l]{\% of responses\\with a signature\\at the end}}}
    & $H_1 \rightarrow B$    & $0\%$ & $0\%$ & $0\%$ & $0\%$ & $0\%$ & $0\%$ & $0\%$\\
    & $H_1 \rightarrow H_2$   & $0\%$ & $100\%$ & $100\%$ & $96.08\%$ & $99.40\%$ & $98.19\%$ & $58.74\%$\\
    & $B \rightarrow H_1$   & $0\%$ & $0\%$ & $0\%$ & $0\%$ & $0\%$ & $0\%$ & $0\%$\\
    & $B \rightarrow H_2$   & $0\%$ & $100\%$ & $100\%$ & $89.73\%$ & $100\%$ & $99.34\%$ & $61.54\%$\\
    \bottomrule
\end{tabular}}
\caption{Effects of style transfer on the observed style properties such as crispness, diversity in vocabulary, and signature at the end of responses (as described in Table \ref{tab:quantitative-style}). Note that, signing names at the end of a response is a unique structural style property of the style $H_2$, hence, this style property is obtained by the models only when transferring a source style to $H_2$. The statistics are obtained on the test set as described in Table \ref{tab:dataset-summary}.}\label{tab:appendix-observed-style-properties}
\end{table*}

\end{document}